%% file: main.tex
\documentclass{article}

\usepackage{PRIMEarxiv}
\usepackage{graphicx}%
\usepackage{multirow}%
\usepackage{amsmath,amssymb,amsfonts}%
\usepackage{amsthm}%
\usepackage{mathrsfs}%
\usepackage[title]{appendix}%
\usepackage{xcolor}%
\usepackage{textcomp}%
\usepackage{manyfoot}%
\usepackage{booktabs}%
\usepackage{algorithm}%
\usepackage{algorithmicx}%
\usepackage{algpseudocode}%
\usepackage{listings}%

\usepackage[numbers]{natbib}
\usepackage{balance}
\usepackage{amsmath,amssymb,amsfonts}
\usepackage{mathtools}
\usepackage{bm}
\usepackage{siunitx}            
\usepackage{booktabs}
\usepackage{tabularx}
\usepackage{multirow}
\usepackage{float}
\usepackage{enumitem}
\usepackage{caption}
\usepackage{subcaption}
\usepackage{xcolor}
\usepackage{placeins}
\usepackage{pgfplots}
\pgfplotsset{compat=1.18}       
\usepgfplotslibrary{groupplots,fillbetween}

\usepackage{tikz}
\usetikzlibrary{patterns,arrows.meta,positioning,calc,fit,shapes.geometric}

\usepackage{multirow}
\usepackage{booktabs}
\usepackage{subcaption}
\usepackage{bm}
\usepgfplotslibrary{colormaps}
\usepackage{tikz}
\usepackage{colortbl}
\usepackage[utf8]{inputenc} 
\usepackage[T1]{fontenc}    
\usepackage{hyperref}       
\usepackage{url}            
\usepackage{booktabs}       
\usepackage{amsfonts}       
\usepackage{nicefrac}       
\usepackage{microtype}      
\usepackage{lipsum}
\usepackage{fancyhdr}       
\usepackage{graphicx}       
\graphicspath{{media/}}     

\title{Joint Domain-Class Modeling for Federated Learning Under Feature Skew
}

\author{
    Sina Najafi \\
	School of Electrical and Computer Engineering \\
	College of Engineering, University of Tehran \\
	Tehran, Iran \\
	\texttt{sinanajafi7878@ut.ac.ir}
	\And
	Mostafa Tavassolipour \\
	School of Electrical and Computer Engineering \\
	College of Engineering, University of Tehran \\
	Tehran, Iran \\
	\texttt{tavassolipour@ut.ac.ir}
	\And
	Seyed Pooya Shariatpanahi \\
	School of Electrical and Computer Engineering \\
	College of Engineering, University of Tehran \\
	Tehran, Iran \\
	\texttt{p.shariatpanahi@ut.ac.ir}
}

\begin{document}
	\maketitle

	\begin{abstract}
		Federated learning (FL) enables collaborative model training without centralizing
		private data, but performance often degrades under \emph{feature skew}: clients
		share labels while the conditional input distributions \(p_i(x\!\mid\!y)\) vary
		due to latent, client-specific appearance factors. We propose \textbf{J}oint
		\textbf{D}omain--Class \textbf{F}ederated \textbf{L}earning (\textbf{JDFL}), a lightweight, optimizer-agnostic
		extension that makes this latent domain variation usable without sharing raw
		data. JDFL first infers \emph{domain clusters} called pseudo-domains from brief local update
		signals. It then expands the classifier head to output $M\times C$, joint
		(domain-class) logits. This allows the model to represent domain-conditioned appearance
		while keeping a shared backbone. To train the expanded head we introduce two
		complementary supervision strategies based on simple intuitions: a
		similarity-aware soft-labeling that transfers evidence between nearby inferred
		domains while allowing domain-specific specialization, and a per-sample
		randomized target assignment that perturbs supervision across the joint outputs
		and serves as a low-cost training-time regularizer. JDFL integrates with
		existing standard FL methods (e.g., FedAvg, SCAFFOLD) with minimal changes.
		Empirically, both supervision modes consistently improve global test accuracy
		on standard domain-shifted image benchmarks; ablations and sensitivity studies show the gains stem from the
		proposed supervision and parametrization rather than mere capacity increase.
	\end{abstract}


	\section{Introduction}
	Federated learning (FL) enables collaborative model training across many edge
	clients while keeping raw data local, which makes it an attractive paradigm
	for privacy-sensitive applications \cite{mcmahan2017communication}. It is used in
	domains such as healthcare \cite{li2025from,qayyum2022collaborative,haripriya2026flipmed,makki2026explainable},
	IoT \cite{rey2022federated,kuang2026brite,ARSALAN2026Digital,XU2026two}, and
	recommendation systems \cite{hard2018federated,hartmann2019federated,Badrouni2026}.
	
	A major challenge for FL is \emph{feature skew} (domain shift): clients share
	labels but their conditional input distributions $p_i(x\mid y)$ differ due to
	client-specific appearance factors (imaging style, sensor characteristics,
	rendering, etc.). In a federated setting with non-iid data, we assume data
	generation at each client is influenced by two approximately independent
	factors: the semantic class $y$ and a client-specific domain variable $d$; thus
	the conditional appearance is more precisely $p(x\mid y,d)$. Forcing a single global
	predictor $p(y\mid x)$ to explain all clients can cause optimization conflict,
	slow convergence and degraded generalization because one parameter vector must reconcile multiple domain-conditioned renderings of the same class.
	
\begin{figure*}[t]
		\centering
		\IfFileExists{overview.pdf}{%
			\includegraphics[page=1,width=\textwidth]{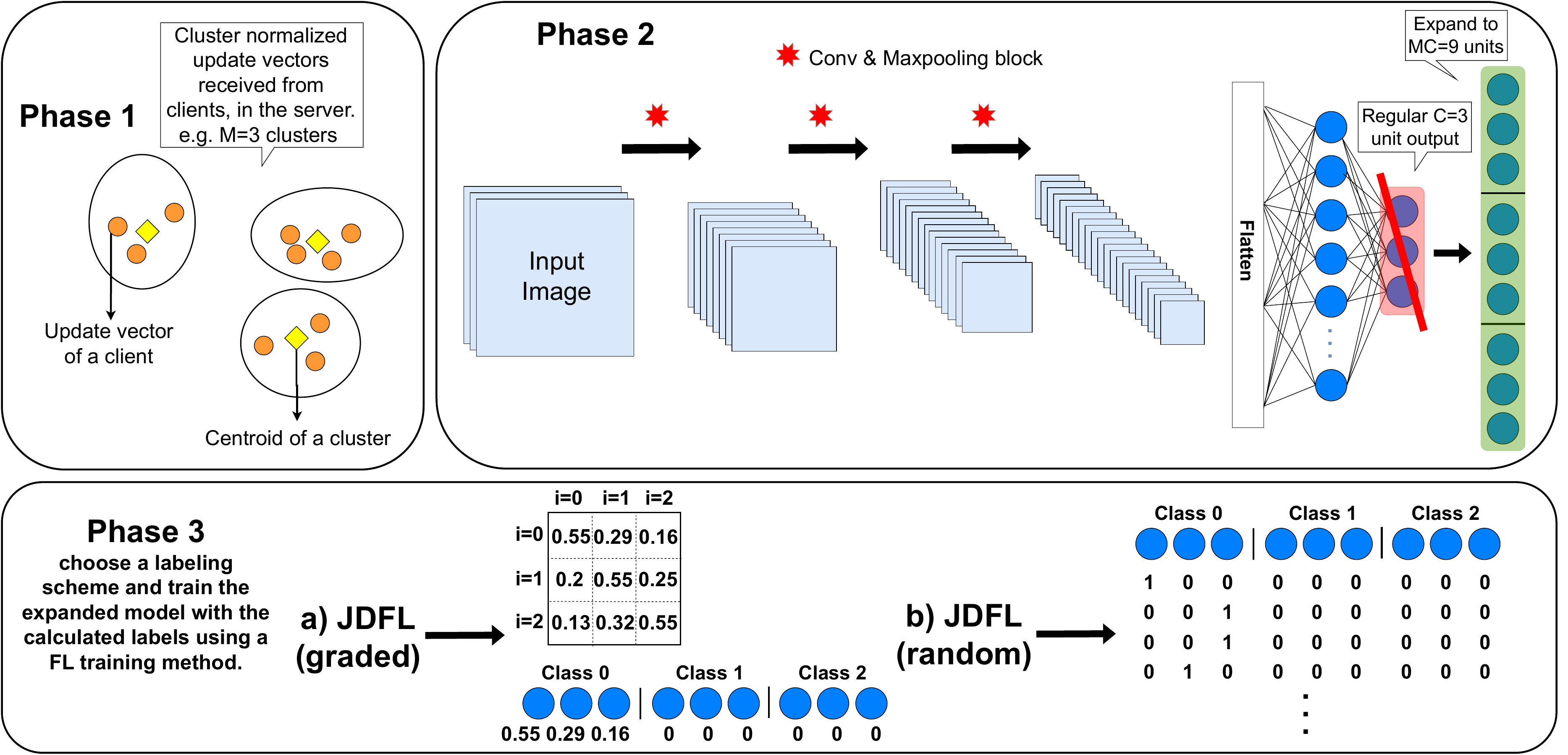}%
		}{%
			\fbox{\parbox{\textwidth}{\centering [Figure missing: overview.pdf]}}%
		}
		\caption{\textbf{Overview of Joint Domain-Class Federated Learning (JDFL).}
			Phase~1: clients run brief local updates and send normalized update vectors to the server,
			which clusters them to infer $M$ domain groups and centroids. 
			Phase~2: the global model expands its classifier head to $M\times C$ joint (domain-class) outputs.
			Phase~3: training proceeds with one of two supervision schemes: 
			(\emph{graded}) similarity-weighted soft targets that share information between related domains, or 
			(\emph{random}) per-sample randomized targets that act as a lightweight regularizer.
			An example for class~0 and domain~0 is shown for both schemes.}
		\label{fig:overview}
	\end{figure*}

	We propose \emph{Joint Domain--Class Federated Learning} (JDFL), a lightweight
	and optimizer-agnostic framework that reconciles domain-specific specialization
	with cross-domain sharing. As illustrated in Fig.~\ref{fig:overview}, JDFL
	operates in three phases:
	\begin{enumerate}
		\item \textbf{Domain discovery:} a brief discovery phase makes latent domain
		structure approximately observable. Clients run a few local steps from a
		shared initialization; the server clusters their normalized update
		vectors to infer $M$ \emph{pseudo-domains} and centroids
		$\{\mu_d\}_{d=1}^M$.
		\item \textbf{Model expansion:} the classifier head is expanded from $C$
		outputs to $M\times C$ joint (pseudo-domain, class) logits, allowing
		the model to represent domain-conditioned appearance while sharing a
		common backbone.
		\item \textbf{Flexible supervision:} training proceeds with supervision on
		the expanded head that trades per-domain specialization and cross-domain
		sharing.
	\end{enumerate}
	
	To balance specialization and transfer we propose two complementary
	supervision schemes. The \emph{graded-invariance} scheme constructs similarity-aware soft targets using centroid cosine-similarities: for a sample of class
	$c$ from a client assigned to pseudo-domain $d$ most supervisory mass is placed
	on the joint unit $j(d,c)$ (the output unit corresponding to class $c$ and pseudo-domain $d$) while the remaining mass is distributed across the
	same-class joint units $j(d',c)$ in other pseudo-domains proportionally to
	their centroid similarity to $d$. This graded soft labeling yields principled,
	similarity-weighted transfer and is explicitly robust to clustering errors.
	
	The \emph{randomized joint-label} scheme acts as a stochastic target perturbation
	rather than an attempt to recover true domain labels. For a sample with true
	class~$c$, the method randomly selects a pseudo-domain $d\in\{1,\dots,M\}$ and
	assigns the primary supervisory mass to the corresponding joint output~$j(d,c)$,
	ensuring that supervision always targets the correct class while randomizing its
	domain-specific instantiation. Over many updates, this stochastic supervision
	encourages the affine scorers in the expanded head to become mutually compatible
	and interchangeable. By injecting controlled randomness into the target space,
	it serves as an effective training-time regularization mechanism that reduces
	over-reliance on any single joint output and helps the optimizer escape shallow
	local minima.

	Empirically, both schemes improve global (class-only) accuracy; the randomized
	variant often yields stronger average gains, while graded-invariance is
	preferable when reliable inter-domain similarity information is available and
	some degree of per-domain specialization is desired.
	
	We validate JDFL across multiple federated optimizers and domain-shifted image
	benchmarks. In addition to standard evaluations, we perform targeted analyses
	including an expanded-head ablation to rule out capacity-only explanations
	and sensitivity sweeps over $M$ to characterize robustness to the number of
	inferred pseudo-domains.
	
	To summarize, our contributions are:
	\begin{itemize}
		\item JDFL: a compact, optimizer-agnostic framework that uncovers pseudo-domains
		from brief local updates and converts feature skew into a structured
		$M\times C$ joint-label problem compatible with standard FL pipelines.
		\item \emph{Graded-invariance}: a cosine-weighted soft-labeling scheme that
		enables similarity-aware transfer between inferred pseudo-domains and is
		robust to clustering noise.
		\item \emph{Randomized joint-label}: a simple per-sample regularizer that
		perturbs supervision across the expanded head and often improves
		class-level generalization.
		\item Extensive empirical validation across federated optimizers, ablations,
		and sensitivity analyses demonstrating the practical benefits and trade-offs.
	\end{itemize}
	
	\section{Related Work}
	As our proposed method is designed as a lightweight extension that can be integrated with a variety of base optimizers, this section provides a broad overview of federated learning optimization techniques with a focus on addressing the data heterogeneity challenge, organized into five principal categories.
	
	\subsection{Client Update Correction Methods}
	
	In non-IID federated learning, the primary issue is \emph{client drift}: local models trained on heterogeneous data tend to deviate from the global objective. The standard FedAvg \cite{mcmahan2017communication} algorithm, which averages these divergent updates, can consequently suffer from slow or unstable convergence. Client update correction methods aim to mitigate this issue by explicitly compensating for the drift. For instance, SCAFFOLD \cite{karimireddy2020scaffold} introduces control variates on both the server and clients to estimate the client drift and correct local updates accordingly.
	
	\subsection{Local Regularization Methods}
	
	These methods counteract client drift by introducing a regularization term into the local objective, encouraging consistency between local and global updates. A foundational method, FedProx \cite{li2020federated}, adds a proximal term that penalizes the Euclidean distance between the local model and the received global model parameters. Other approaches operate at the feature level. For example, MOON \cite{li2021model} utilizes a contrastive loss to maximize the agreement between representations learned by the current local model and the previous global model, encouraging consistency in the learned feature space. FedDC \cite{gao2022feddc} also uses an auxiliary local variable in its regularization term to track and correct for the historical gap between the local and global models. FedDyn \cite{acar2021federated} dynamically adds a regularization term that uses a persistent server-side auxiliary variable and per-client correction variables to steer each client's local optimum toward the global objective. More recently, FedProc \cite{mu2023fedproc} proposes a prototype-guided framework that aligns local representations with global class prototypes. By enforcing this prototypical consistency during training, FedProc directly reduces the divergence of feature distributions across clients, thereby improving generalization under heterogeneous data.
	
	\subsection{Knowledge Distillation Methods}
	
	Knowledge distillation is used in FL for model aggregation, where client models act as an ensemble of "teachers" to train a global "student" model on the server by aligning their output logits \cite{hinton2015distilling, li2019fedmd}. Early federated distillation methods like FedDF \cite{lin2020ensemble} required a public, unlabeled dataset on the server to facilitate this knowledge transfer. However, the practical limitation of needing such a proxy dataset has led to the development of data-free approaches. More recent works, such as FedGen \cite{zhu2021datafree}, overcome this issue by learning a generative model on the server. This generator creates synthetic data samples that are then used to distill knowledge from the client models into the global model, removing the dependency on any external dataset.
	
	\subsection{Data Sharing Methods}
	
	Data sharing methods address heterogeneity by directly augmenting the clients' local datasets. The motivation is to provide clients with samples from under-represented or absent classes to create a more balanced training distribution. Common strategies include sharing a common public dataset \cite{zhao2018federated}, distributing centrally synthesized data \cite{hao2021towards}, or exchanging a small subset of clients' real training samples \cite{tuor2020overcoming}. While effective at mitigating the non-IID problem, this entire category introduces significant privacy concerns. Sharing any form of raw or artificial input data is in direct tension with the core privacy-preserving principles of Federated Learning.
	
	\subsection{Personalized FL Methods}
	
	In contrast to methods that train a single global model, personalized federated learning (PFL) aims to build a customized model for each client, which is often more effective under significant data heterogeneity. One line of research frames this challenge through the lens of meta-learning or multi-task learning, treating each client as a distinct task and learning a global model that can be easily adapted locally \cite{fallah2020personalized,jiang2019improving,chen2018federated,smith2017federated}. Another popular approach involves creating hybrid models with both shared global parameters and private local parameters. The global components learn from all clients, while the private components are trained only on local data to specialize the model \cite{arivazhagan2019federated,collins2021exploiting,bui2019federated,oh2022fedbabu}. A third strategy involves clustering clients based on the similarity of their data or models and then performing federated aggregation only within each cluster, resulting in a distinct model for each client group \cite{ghosh2019robust,ghosh2022efficient,long2023multicenter}.
	
	In this work, we consider training a single global classification model. We show that adding our extension to the existing compatible methods will lead to consistent improvement in accuracy.
	We emphasize that although JDFL does an initial clustering, it is an \emph{extension} rather than an independent clustered FL algorithm. Basically JDFL is orthogonal to the choice of optimizer or clustering/personalization strategy: it is intended to \emph{augment} existing methods by providing a lightweight method.
	
	\section{Problem setup and notation}
	
	We consider a federated learning setting with a set of clients
	\(\mathcal{S}=\{1,\dots,N\}\), where \(N\) denotes the total number of clients.
	Each client \(i\in\mathcal{S}\) holds a private labeled dataset
	\(\mathcal{D}_i=\{(x_{i,n},y_{i,n})\}_{n=1}^{n_i}\) with \(x_{i,n}\in\mathcal{X}\) and
	semantic labels \(y_{i,n}\in\{1,\dots,C\}\).
	
	The heterogeneity targeted in this work is \emph{feature skew} (domain
	heterogeneity): clients originate from different data-generating domains so that
	conditional input distributions \(p_i(x\mid y)\) vary across clients. Multiple
	clients may share the same domain. ground-truth domain labels are unavailable
	to the server.
	
	Let \(M\) denote the number of \emph{pseudo-domains} discovered by a clustering
	step and write \(\mathcal{M}=\{1,\dots,M\}\). We consider a classifier
	\(f(\cdot;\theta):\mathcal{X}\to\mathbb{R}^{M C}\) whose final linear head
	produces \(M C\) logits. For domain index \(d\in\mathcal{M}\) and class
	\(c\in\{1,\dots,C\}\) we define the flattened joint index
	\begin{equation}\label{eq:joint_index}
		j(d,c) = (d-1)\cdot C + c,
	\end{equation}
	so that \(j(d,c)\in\{1,\dots,MC\}\).
	
	Primary evaluation focuses on \emph{class-only accuracy}: if the model
	predicts joint index \(\hat j\), the corresponding class prediction is
	\(\hat c = ((\hat j-1)\bmod C) + 1\). In other words the model is correct
	if the largest of the \(MC\) logits belongs to any of the \(M\) outputs that
	correspond to the true class (When \(M=1\) this reduces to ordinary accuracy). 
	
	We use the following shorthand for algorithmic hyperparameters:
	\(E_0\) (short local epochs in the discovery phase), \(E\) (local epochs in
	main training), \(T\) (communication rounds), \(\eta_\ell\) (local step size),
	\(\tau\) (softmax temperature for cosine weights), and \(\alpha\) (main-label
	probability used in the graded invariance soft targets). For the random-assignment
	variant we denote the chosen-unit mass by \(\phi\in[0,1]\) (so
	\(\phi=1\) recovers a one-hot/hard assignment on the chosen output).

	\section{Method}
	
	\subsection{Motivation and high-level design}
	
	As discussed in the introduction, clients' conditional input distributions differ
	because data from each client is influenced by class and a client-specific domain factor.
	This produces domain-conditioned renderings of the same class and can induce gradient
	conflict, slower convergence, and degraded generalization when a single global predictor
	must reconcile all such variations. This observation motivates our two complementary designs: one that explicitly trades per-domain specialization against cross-domain sharing, and another that injects stochasticity into supervision to help escape insignificant local minima gaining better generalization.

	
	Our framework follows three design choices motivated by this generative view:
	\begin{enumerate}
		\item make the latent domain structure approximately observable via
		clustering of brief local update vectors,
		\item expand the classifier to explicitly represent the joint target
		\((Y,D)\) so the model can capture domain-conditioned appearance and
		\item introduce two complementary supervision schemes on the expanded head
		that trade specialization and invariance in different ways:
		\begin{itemize}
			\item \textbf{Graded invariance (clustering + cosine soft labels)}:
			principled, similarity-weighted sharing between inferred
			pseudo-domains (robust to clustering noise).
			\item \textbf{Random assignment (sample-wise randomized targets)}:
			a simple, model-driven regularizer that leverages the final-classifier
			nonlinearity by randomly assigning which joint output receives the
			primary supervisory mass on a per-sample basis. This perturbation
			combined with the non-linear (piecewise linear) classifier acts as label regularization and can improve optimization and
			generalization in practice.
		\end{itemize}
	\end{enumerate}
	
	Below we describe the two supervision schemes formally, explain how they
	integrate with federated optimizers,
	and provide practical details for pseudo-domain discovery and evaluation.
	
	\subsection{Phase 1: pseudo-domain discovery}
	
	Brief local updates from a common initialization reveal
	how the shared backbone adapts to each client's feature distribution; clustering
	these signals groups clients by similar domain biases.
	
	\paragraph{Update vectors and normalization.} Let \(\theta_0\) be the shared
	initialization. Each client \(i\) performs brief local training for \(E_0\)
	epochs producing updated parameters \(\tilde\theta_i\). For a chosen set of
	high-level parameters (layers) \(\mathcal{P}\) (e.g., the last convolutional block and the
	classification head) we define
	\begin{equation}\label{eq:update_vector}
		u_i = \operatorname{vec}\big(\tilde\theta_i[\mathcal{P}] - \theta_0[\mathcal{P}]\big)\in\mathbb{R}^D,
	\end{equation}
	where \(\operatorname{vec}(\cdot\)) flattens and concatenates the selected tensors.
	Normalize \(u_i\) as
	\begin{equation}\label{eq:update_normalize}
		\bar u_i = \frac{u_i}{\|u_i\|_2}
	\end{equation}
	to reduce scale effects.
	
	\paragraph{Clustering.} Run K-means \cite{arthur2007kmeanspp, lloyd1982least} (or an alternative clustering algorithm)
	on \(\{\bar u_i\}_{i=1}^N\) to obtain assignments \(z_i\in\mathcal{M}\) and
	centroid vectors \(\mu_1,\dots,\mu_M\). Choose \(M\) via elbow/silhouette
	criteria. We explicitly treat \(z_i\) as \emph{pseudo}-domain
	labels (not domain labels) because clustering can be noisy; the supervised targets in Phase 2 are
	designed to be robust to misassignments.
	
	\subsection{Phase 2: joint-label parametrization}
	
	\paragraph{Joint parametrization.}
	After discovery, adapt the classifier so the final head outputs \(M C\) logits
	indexed by \(j(d,c)\) from Eq.~\eqref{eq:joint_index}. This lets the model
	allocate capacity per (class, pseudo-domain) pair while sharing a common
	backbone. When evaluated with class-only accuracy, allocating \(M\) outputs
	per class makes each class score the pointwise maximum of \(M\) affine functions,
	producing a non-linear (and thus more expressive) decision function. by
	contrast, a single output per class implements only one affine score (The proof of non-linearity (piecewise linearity) of the expanded classifier head can be found in Appendix~\ref{app:A}). A larger number of pseudo-domains \(M\) can therefore improve the expressive
	power of the classifier head, but overly large \(M\) inflates the output layer
	and may lead to overfitting or increased optimization difficulty, thereby
	diminishing the potential benefits. To balance these factors, we can cap the number of pseudo-domains
	at some threshold (e.g. 10).
	
	Below we present two distinct but related supervision schemes that operate on
	the same expanded head.
	
	\subsubsection{(A) Graded domain invariance via centroid-weighted soft labels}
	This is the method that combines a notion of domain invariance (graded domain invariance) with federated learning.
	
	For a client assigned to pseudo-domain \(d\) and a sample of class \(c\) we
	build a target probability vector \(t^{(d,c)}\in\Delta^{MC}\) as follows.
	
	\begin{enumerate}
		\item Compute cosine similarities between centroid \(\mu_d\) and other centroids \(\mu_{d'}\):
		\begin{equation}\label{eq:cosine_sim}
			s_{d,d'} = \frac{\mu_d^\top\mu_{d'}}{\|\mu_d\|_2\|\mu_{d'}\|_2}\quad (d'\neq d).
		\end{equation}
		\item Convert these to normalized weights (temperature \(\tau>0\)):
		\begin{equation}\label{eq:soft_weights}
			\tilde w_{d'}=\frac{\exp(s_{d,d'}/\tau)}{\sum_{r\neq d}\exp(s_{d,r}/\tau)}\quad (d'\neq d).
		\end{equation}
		\item Place mass \(\alpha\) on the local joint index and distribute \(1-\alpha\)
		across the same class in other pseudo-domains:
		\begin{equation}\label{eq:graded_target}
			t^{(d,c)}_{\,j(d,c)}=\alpha,\;
			t^{(d,c)}_{\,j(d',c)}=(1-\alpha)\tilde w_{d'}\;(\forall\, d'\neq d).
		\end{equation}
		
		and all other entries are zero. This guarantees \(\sum_j t^{(d,c)}_j=1\).
	\end{enumerate}
	
	The resulting supervised objective for a sample \((x,c)\) from pseudo-domain
	\(d\) is the KL / cross-entropy between \(t^{(d,c)}\) and the model softmax
	distribution \(p=\operatorname{softmax}(f_\theta(x))\):
	\begin{equation}\label{eq:loss_graded}
		\mathcal{L}_{\text{graded}}(\theta;x,c,d) = -\sum_{j=1}^{MC} t^{(d,c)}_j\log p_j.
	\end{equation}
	
	\paragraph{Intuition and trade-offs.} The parameter \(\alpha\in(0,1]\) controls
	a tradeoff between local specialization (large \(\alpha\))
	and cross-domain sharing (small \(\alpha\)). Because residual mass is distributed
	proportionally to centroid similarity, the supervision is \emph{graded}:
	similar pseudo-domains are encouraged to share more; dissimilar pseudo-domains transfer
	less knwoledge and therefore specialize more. Graded invariance reduces harmful gradient conflict while allowing
	domain-conditioned specialization where necessary; it is explicitly robust to
	clustering noise because targets are never strictly committed to a single
	(possibly incorrect) pseudo-domain.
	A concrete toy example illustrates the idea: with \(M = 4\), \(\alpha = 0.45\), and centroid similarities that strongly favor domain 2, a plausible soft target for class \(c\) is:

	\[
	t^{(d,c)} = [0.45,\ 0.39,\ 0.09,\ 0.07],
	\]

	i.e., 45\% mass on the local domain\(\times\)class logit, 39\% on a very similar domain, and only a tiny mass on distant domains. This tells the model: "be fairly confident this sample belongs to class \(c\) in domain \(d\), but allow non-negligible ambiguity toward domain 2 (because they are similar)."

	\subsubsection{(B) Randomized joint-label assignment (per-sample)}
	
	This complementary scheme does not attempt to recover domain labels. Instead, it randomly perturbs supervision on the expanded head to exploit two practical effects: (i) the non-linear, piecewise structure of the final classifier (an \(M\)-fold expansion per class) which allows different joint outputs for the same class to act as interchangeable affine scorers, and (ii) stochastic target perturbation which functions as a strong regularizer and can help the optimizer escape insignificant local minima.
	
	Concretely, in the \emph{per-sample} mode each training sample independently draws a chosen pseudo-domain index
	\[
	d^\star \sim \operatorname{Uniform}(\{1,\dots,M\}),
	\]
	(resampled per sample and per epoch) and the target over the \(M\) domain outputs for the sample's class \(c\)
	is constructed as in Eq.~\eqref{eq:random_target}. The parameter \(\phi\in[0,1]\) controls how strongly the chosen unit is emphasized.
	
	\begin{equation}\label{eq:random_target}
		\begin{aligned}
			t^{(d^\star,c)}_{\,j(d',c)} &=
			\begin{cases}
				\phi, & d' = d^\star,\\[4pt]
				\dfrac{1-\phi}{M-1}, & d' \neq d^\star,
			\end{cases} \\[6pt] 
			&\hspace{-2.5em}\text{and}\quad
			t^{(d^\star,c)}_{\,j(d',c')} = 0 \text{ for } c' \neq c.
		\end{aligned}
	\end{equation}

	\paragraph{Intuition and trade-offs.} Randomized assignment uses the extra degrees of freedom in the expanded head as interchangeable scorers for the same class; by randomly switching which scorer receives the primary supervisory mass, the model learns features that generalize across these scorers. Because the assignments are stochastic, this behaves similarly to label perturbation or label smoothing: it discourages the network from deterministically relying on any single joint output and injects gradient noise that can improve optimization (e.g., escaping shallow minima). Empirically we find the per-sample variant often yields stronger average class-only accuracy, but the downside is reduced per-domain specialization compared to the graded scheme: randomized training encourages invariance-like behaviour at the cost of tailored domain specialization.
	
	\subsection{Phase 3: federated optimization and Loss function}
	
	\paragraph{Integration with federated optimizers.} The joint-label objective is
	optimizer-agnostic and can be combined with a range of federated optimizers. after the discovery and head-expansion steps the method requires only two local changes in the standard FL loop. First, clients construct per-sample joint targets for the expanded $M\times C$ head according to either the graded or randomized rule; second, local updates are computed by minimizing the joint-head objective and then returned to the server for aggregation. All server-side logic (client selection, aggregation and global model update) remains unchanged, so JDFL plugs into many existing federated optimizers with minimal engineering effort.
	Algorithm~\ref{alg:jdfl_compact_ref} presents the three phases of JDFL. 
	
	\paragraph{Loss.} Both supervision schemes (graded and random) minimize the divergence between the
	target distribution \(t\) and model prediction
	\(p=\operatorname{softmax}(f_\theta(x))\) via
	\(\mathcal{L}_{\text{soft}}=\operatorname{KL}(t\,\|\,p)\), which generalizes
	cross-entropy when \(t\) is one-hot. \newline
	\newline
	\newline

	\begin{algorithm}[t]
		\caption{Joint Domain-Class Federated Learning}
		\label{alg:jdfl_compact_ref}
		\centering
		\small
		\begin{minipage}{0.95\columnwidth}
			\begin{algorithmic}[1]
				\Require Clients \(1,\dots,N\) with datasets \(\mathcal{D}_i\), init.\ \(\theta_0\);
				clustering layers \(\mathcal{P}\); discovery epochs \(E_0\); chosen \(M\);
				federated rounds \(T\); local epochs \(E\); target params.
				\Ensure final \(\theta_T\), assignments \(z_i\), centroids \(\{\mu_d\}\).
				
				\State \textbf{Phase 1 (Discovery).} Each client \(i\) runs \(E_0\) local epochs
				from \(\theta_0\), computes the update vector \(u_i\) (Eq.~\ref{eq:update_vector}),
				normalizes \(\bar u_i\) (Eq.~\ref{eq:update_normalize}), and sends it to the server.
				\State Server clusters \(\{\bar u_i\}\) into \(M\) pseudo-domains, producing assignments \(z_i\) and centroids \(\{\mu_d\}\).
				\State \textbf{Phase 2 (Parametrization \& targets).} Expand head to \(M\times C\) logits (indexing by Eq.~\ref{eq:joint_index}).
				For each client sample construct targets either by graded rule (Eq.~\ref{eq:graded_target}) or random rule (Eq.~\ref{eq:random_target}).
				\State \textbf{Phase 3 (Federated training).} For rounds \(t=1,\dots,T\):
				\State \quad Select clients \(S_t\). Each \(i\in S_t\) pulls \(\theta_{t-1}\), runs \(E\) local epochs minimizing the per-sample loss (KL to the target; graded loss in Eq.~\ref{eq:loss_graded} when applicable), and returns updates. Server aggregates using the chosen federated optimizer.
				\State \Return \(\theta_T, \{z_i\}, \{\mu_d\}\).
			\end{algorithmic}
		\end{minipage}
		\normalsize
	\end{algorithm}

	\section{Experiments}
	
	Our experiments aim to (i) validate that JDFL improves class-level generalization under domain heterogeneity, and (ii) verify compatibility with existing federated optimizers (FedAvg, SCAFFOLD, FedDyn, FedDC, FedProc).
	
	\subsection{Datasets and data partitioning}
	\paragraph{PACS.} PACS (Photo, Art painting, Cartoon, Sketch) is used as a canonical domain-shift benchmark. For PACS federated experiments we partitioned the dataset across 10 and 40 clients. Three federated settings are reported:
	\begin{itemize}
		\item \textbf{Setting A (fewer participants, many small local updates):} local epochs $E=1$, 10 clients, client fraction $=1.0$ (all clients participate each round).
		\item \textbf{Setting B (fewer participants, longer local work):} local epochs $E=5$, 10 clients, client fraction $=0.5$ (half the clients sampled each round).
		\item \textbf{Setting C (many participants):} local epochs $E=3$, 40 clients, client fraction $=0.5$ (half the clients sampled each round).
	\end{itemize}
	In each setting data were split so that all domains appear across the clients but each client has data of a single domain (which is a reasonable assumption). Each reported result is averaged across 3 random seeds.
	
	\paragraph{Office--Caltech.} For Office--Caltech we focus on the scenario with \emph{data-size heterogeneity} across domains. Because some domains have few examples, we assign each \emph{domain} to a single client (so the number of clients equals the number of domains in the dataset). We report results for the federated setting with local\_epochs $E=5$ and client fraction $=1.0$ to evaluate the method under extreme domain imbalance.
	
	\subsection{Model architecture}
	We use a lightweight CNN model for our classification task. Concretely, the backbone is a 3-block convolutional network with GroupNorm, followed by two hidden fully connected layers and a final linear classification head. The input shape to our model is $3\times112\times112$. The model architecture detail is available in Appendix~\ref{app:B}. In the joint-label experiments the final head is expanded to $M\times C$ logits.
	In preliminary experiments we also evaluated standard backbones (e.g. ResNet-18 with random initialization). On our small domain-generalization benchmarks these larger networks consistently overfit and produced worse held-out performance than the shallow model in the baseline experiments, which is expected given the limited per-domain sample sizes; for this reason we use the lightweight CNN throughout. Importantly, every baseline and its JDFL-augmented variant share the same backbone, so reported gains are attributable to the $M\times C$ parametrization and supervision schemes rather than differences in backbone capacity. Evaluating JDFL on larger datasets that can be used with larger models is left to future work.
	
	\setlength{\tabcolsep}{2pt}        

	\begin{table}[t]
		\centering
		\small                          
		\setlength{\tabcolsep}{2.5pt}   
		\renewcommand{\arraystretch}{1.2}
		\caption{Standalone vs.\ Baseline+JDFL for Office-Caltech, PACS settings A, B, and C. Numbers represent the mean overall class-only global test accuracy across 3 seeds.}
		\label{tab:all_exp}
		\begin{tabular*}{\columnwidth}{@{\extracolsep{\fill}}|l|c|c|c|c|}
			\hline
			Method & Acc Office & Acc PACS(A) & Acc PACS(B) & Acc PACS(C) \\
			\hline
			FedAvg            & 73.10 & 69.40 & 67.90 & 68.53 \\
			+JDFL graded      & 76.51 ({\textcolor{green}{+3.41}}) & 70.30 ({\textcolor{green}{+0.90}}) & 69.87 ({\textcolor{green}{+1.97}})  & 69.23 ({\textcolor{green}{+0.70}}) \\
			+JDFL random      & \textbf{76.97} ({\textcolor{green}{+3.87}}) & \textbf{71.80} ({\textcolor{green}{+2.40}}) & \textbf{70.53} ({\textcolor{green}{+2.63}}) & \textbf{70.07} ({\textcolor{green}{+1.54}}) \\
			\hline
			SCAFFOLD          & 75.20 & 70.63 & 69.83 & 69.27 \\
			+JDFL graded      & 77.37 ({\textcolor{green}{+2.17}}) & \textbf{71.97} ({\textcolor{green}{+1.34}}) & 71.83 ({\textcolor{green}{+2.00}}) & \textbf{70.47} ({\textcolor{green}{+1.20}}) \\
			+JDFL random      & \textbf{78.61} ({\textcolor{green}{+3.41}}) & 71.86 ({\textcolor{green}{+1.23}}) & \textbf{72.80} ({\textcolor{green}{+2.97}}) & 70.13 ({\textcolor{green}{+0.86}}) \\
			\hline
			FedDyn            & 71.72 & 69.00 & 67.47 & 67.43 \\
			+JDFL graded      & 73.10 ({\textcolor{green}{+1.38}}) & \textbf{70.63} ({\textcolor{green}{+1.63}}) & \textbf{68.40} ({\textcolor{green}{+0.93}}) & \textbf{69.50} ({\textcolor{green}{+2.07}}) \\
			+JDFL random      & \textbf{73.88} ({\textcolor{green}{+2.16}}) & 69.57 ({\textcolor{green}{+0.57}}) & 67.60 ({\textcolor{green}{+0.13}}) & 68.67 ({\textcolor{green}{+1.24}}) \\
			\hline
			FedDC             & 74.35 & 68.27 & 69.80 & 69.30 \\
			+JDFL graded      & 75.78 ({\textcolor{green}{+1.43}}) & \textbf{70.50} ({\textcolor{green}{+2.23}}) & \textbf{71.43} ({\textcolor{green}{+1.63}}) & \textbf{69.77} ({\textcolor{green}{+0.47}}) \\
			+JDFL random      & \textbf{77.29} ({\textcolor{green}{+2.94}}) & 69.27 ({\textcolor{green}{+1.00}}) & 70.03 ({\textcolor{green}{+0.23}}) & 69.37 ({\textcolor{green}{+0.07}}) \\
			\hline
			FedProc           & 76.82 & 71.03 & 69.03 & 69.07 \\
			+JDFL graded      & \textbf{77.95} ({\textcolor{green}{+1.13}}) & 72.47 ({\textcolor{green}{+1.44}}) & \textbf{70.57} ({\textcolor{green}{+1.54}}) & 69.73 ({\textcolor{green}{+0.66}}) \\
			+JDFL random      & 77.82 ({\textcolor{green}{+1.00}}) & \textbf{72.57} ({\textcolor{green}{+1.54}}) & 70.17 ({\textcolor{green}{+1.14}}) & \textbf{69.77} ({\textcolor{green}{+0.70}}) \\
			\hline
		\end{tabular*}
	\end{table}

	\subsection{Baselines and algorithmic variants}
	We evaluate the following methods as baselines and as \emph{augmented} variants with JDFL:
	\textbf{FedAvg}, \textbf{SCAFFOLD}, \textbf{FedDyn}, \textbf{FedDC}, \textbf{FedProc}.
	
	For each baseline we report:
	\begin{enumerate}
		\item \textbf{Standalone:} baseline trained as usual (original output head of size $C$). Reported metric: \emph{standard accuracy} (joint head not used).
		\item \textbf{Baseline + JDFL (graded and random):} baseline integrated and trained with our joint-label parametrization and specific labeling schemes. Hyperparameters are discussed in section \ref{sec:hyper} . Reported metric: \emph{class-only accuracy} (defined in section \ref{sec:eval_metrics}).
	\end{enumerate}
	
	\subsection{Compatibility and comparison rationale}
	JDFL is designed to be plug-in: its discovery, expanded-head parametrization, and supervision schemes can be combined with a wide range of federated optimizers or with cluster/personalization pipelines. For a fair and clear evaluation we therefore (i) treat standard optimizers (FedAvg, SCAFFOLD, FedDyn, FedDC, FedProc) as base methods and report results both with and without the JDFL extension, and (ii) measure improvements attributable to the JDFL parametrization/supervision rather than changes in aggregation. Comparing JDFL as a standalone method to other independent single-model, clustered or personalized algorithms would be misleading because although some baseline methods show more significant gains when combined with JDFL, there is no single baseline that JDFL can be combined with; instead we aim to show that adding JDFL to compatible pipelines consistently improves class-level generalization.

	\subsection{Hyperparameters and implementation details}
	\label{sec:hyper}
	Unless otherwise stated, experiments share the following common configuration:
	\begin{itemize}
		\item \textbf{Optimizer (local):} SGD with learning rate $0.01$, no weight decay.
		\item \textbf{Soft-label parameters:} graded JDFL: main probability $\alpha=0.55$ if $M\le5$ and $\alpha=0.45$ if $M>5$, temperature $\tau=0.1$ (we do not extensively tune these). Random JDFL: main probability $\phi=1.0$.
		\item \textbf{Discovery phase:} short local training with $E_0=3$ epochs (used to compute client update vectors for clustering).
		\item \textbf{Clustering:} K-means applied to $\ell_2$-normalized update vectors from the last convolutional block + classification head. The number of pseudo-domains $M$ is selected via elbow inspection.
		\item \textbf{Other hyper-parameters:} standalone baselines and their JDFL-augmented counterparts use the same shared hyper-parameter values (details in Appendix~\ref{app:C}).
	\end{itemize}
	
	\subsection{Evaluation metrics}
	\label{sec:eval_metrics}
	We report the following evaluation metrics.
	
	\paragraph{Standard accuracy.} For standalone baselines (no joint head) we report the usual test accuracy: fraction of examples whose argmax logit equals the ground truth class.
	
	\paragraph{Class-only accuracy.} When a joint head of size $M\times C$ is used (baseline+JDFL), each model prediction is a joint index $\hat j\in\{1,\dots,MC\}$. The \emph{class-only} prediction is derived by marginalizing out the pseudo-domain index:
	\[
	\hat c = ((\hat j-1)\bmod C) + 1.
	\]
	Class-only accuracy is measured as the fraction of test examples for which $\hat c$ equals the ground truth class. This metric isolates semantic (class) performance from the pseudo-domain conditioning and is the primary metric for evaluating JDFL combinations. Standard accuracy equals class-only accuracy when $M=1$.
	
	\paragraph{Per-domain accuracy.} In addition to overall class-only / standard accuracy we report accuracy on each domain separately in appendix~\ref{app:C} (standard accuracy for standalone experiments and class-only accuracy for joint experiments).
	
	Each client holds out a portion of its dataset for testing; we report metrics on the union of these test sets.
	
	\subsection{Results}
	The maximum global test accuracies obtained from both the baseline methods and their extended variants incorporating the proposed techniques are presented in Table~\ref{tab:all_exp}. Overall, the addition of either extension consistently improves global test accuracy. More specifically, the random JDFL variant appears more effective when combined with FedAvg and SCAFFOLD, whereas graded JDFL yields better results with FedDyn, FedDC and FedProc. Further details including Convergence dynamics and per-domain breakdowns are provided in Appendix~\ref{app:C}.
	
	\subsection{Clustering diagnostics}
	\label{sec:clustering_diag_compact}
	To validate pseudo-domain discovery, we ran a compact diagnostic sweep on PACS. We vary discovery epochs $E_0\in\{1,\dots,5\}$, discovery batch size $b\in\{16,32,64\}$, and three layer subsets: A = head only, B = last conv + head, C = last two convs + head. For each configuration we report two clustering-only metrics: (i) Adjusted Rand Index (ARI) with $M$ fixed to the true domain count ($M=4$) in Fig.~\ref{fig:heatmapfirst}, and (ii) elbow-selection reliability: fraction of runs where the elbow procedure returned $\hat M=4$ in Table~\ref{tab:elbow_reliability_compact}. These diagnostics measure whether brief discovery runs produce domain-like clusters and whether the elbow heuristic recovers the true cluster count.

	\begin{figure*}[t]
		\centering
		\begin{minipage}[c]{0.78\textwidth}
			\centering
			\footnotesize
			\setlength{\tabcolsep}{4pt}          
			\renewcommand{\arraystretch}{1.5}    
			\begin{tabular*}{\linewidth}{@{\extracolsep{\fill}}c*{9}{c}}
				\toprule
				\multirow{2}{*}{$E_0$} & \multicolumn{9}{c}{Batch--Layer} \\
				\cmidrule(lr){2-10}
				& b16-A & b16-B & b16-C & b32-A & b32-B & b32-C & b64-A & b64-B & b64-C \\
				\midrule
				5 & \cellcolor{green!94!red}0.969 & \cellcolor{green!94!red}0.970 & \cellcolor{green!92!red}0.958
				& \cellcolor{green!95!red}0.976 & \cellcolor{green!97!red}0.987 & \cellcolor{green!93!red}0.963
				& \cellcolor{green!41!red}0.703 & \cellcolor{green!68!red}0.841 & \cellcolor{green!55!red}0.773 \\
				4 & \cellcolor{green!95!red}0.973 & \cellcolor{green!86!red}0.930 & \cellcolor{green!91!red}0.957
				& \cellcolor{green!83!red}0.916 & \cellcolor{green!90!red}0.952 & \cellcolor{green!81!red}0.904
				& \cellcolor{green!41!red}0.704 & \cellcolor{green!62!red}0.812 & \cellcolor{green!67!red}0.835 \\
				3 & \cellcolor{green!76!red}0.881 & \cellcolor{green!88!red}0.939 & \cellcolor{green!86!red}0.931
				& \cellcolor{green!56!red}0.779 & \cellcolor{green!64!red}0.820 & \cellcolor{green!70!red}0.850
				& \cellcolor{green!44!red}0.722 & \cellcolor{green!62!red}0.808 & \cellcolor{green!54!red}0.771 \\
				2 & \cellcolor{green!58!red}0.792 & \cellcolor{green!65!red}0.826 & \cellcolor{green!69!red}0.847
				& \cellcolor{green!62!red}0.809 & \cellcolor{green!69!red}0.843 & \cellcolor{green!58!red}0.791
				& \cellcolor{green!51!red}0.757 & \cellcolor{green!49!red}0.745 & \cellcolor{green!53!red}0.763 \\
				1 & \cellcolor{green!35!red}0.676 & \cellcolor{green!37!red}0.687 & \cellcolor{green!41!red}0.704
				& \cellcolor{green!33!red}0.665 & \cellcolor{green!34!red}0.670 & \cellcolor{green!32!red}0.661
				& \cellcolor{green!22!red}0.608 & \cellcolor{green!35!red}0.674 & \cellcolor{green!35!red}0.674 \\
				\bottomrule
			\end{tabular*}
			
			\vspace{8pt}
			\captionsetup{justification=justified, width=0.95\linewidth}
			\caption{\textbf{ARI Metric for the Initial Clustering.}
				Reliability of update vector clustering when the number of clusters matches
				the actual number of domains. The results are shown for different configurations
				(with 40 clients), each tested with 5 different random seeds. It can be inferred
				that as the batch size gets larger more local updates are needed for reliable clustering.}
			\label{fig:heatmapfirst}
		\end{minipage}%
		\hfill
		\begin{minipage}[c]{0.20\textwidth}
			\centering
			\begin{tikzpicture}
				\begin{axis}[
					colorbar,
					colormap={redgreen}{rgb255(0.5)=(255,0,0) rgb255(1.0)=(0,255,0)},
					colorbar style={
						samples=100,
						width=10pt,
						height=3.0cm,
						point meta min=0.5,
						point meta max=1.0,
						ylabel={ARI},
						ylabel style={rotate=-90, anchor=south, font=\footnotesize},
						ticklabel style={font=\footnotesize},
						ytick={0.5,0.75,1.0},
					},
					hide axis,
					]
					\addplot[draw=none] coordinates {(0,0.5) (0,1.0)};
				\end{axis}
			\end{tikzpicture}
		\end{minipage}
	\end{figure*}

	\begin{table}[t]
		\centering
		\small
		\setlength{\tabcolsep}{4pt}          
		\renewcommand{\arraystretch}{1.15}   
		\caption{Elbow selection reliability: fraction of runs (total 5 runs with 5 different seeds) where elbow selected $\hat M=4$. The upper block is for 10 clients, the lower block for 40 clients. As batch size increases we need more epochs to detect the actual number of domains. As clients increase because we have less data in each client more local updates are needed to be able to detect the actual number of domains.}
		\label{tab:elbow_reliability_compact}
		
		\begin{tabular}{@{}l*{9}{c}@{}}
			\toprule
			$E_0$ & \multicolumn{3}{c}{batch = 16} & \multicolumn{3}{c}{batch = 32} & \multicolumn{3}{c}{batch = 64} \\
			\cmidrule(lr){2-4}\cmidrule(lr){5-7}\cmidrule(lr){8-10}
			& A & B & C & A & B & C & A & B & C \\
			\midrule
			\multicolumn{10}{l}{\textbf{10 clients}} \\
			\midrule
			1 & 20\% & 0\% & 20\%  & 0\% & 0\% & 0\% & 0\% & 0\% & 0\% \\
			2 & 60\% & 100\% & 100\% & 20\% & 20\% & 40\% & 0\% & 20\% & 40\% \\
			3 & 80\% & 100\% & 100\% & 80\% & 100\% & 100\% & 40\% & 60\% & 60\% \\
			4 & 60\% & 60\% & 60\%  & 60\% & 80\% & 80\% & 80\% & 100\% & 100\% \\
			5 & 20\% & 40\% & 80\%  & 40\% & 60\% & 100\% & 100\% & 100\% & 100\% \\
			\midrule
			\multicolumn{10}{l}{\textbf{40 clients}} \\
			\midrule
			1 & 0\% & 0\% & 20\%  & 40\% & 20\% & 20\% & 0\% & 0\% & 20\% \\
			2 & 60\% & 40\% & 60\% & 40\% & 40\% & 20\% & 40\% & 20\% & 80\% \\
			3 & 60\% & 80\% & 80\% & 60\% & 20\% & 20\% & 20\% & 40\% & 20\% \\
			4 & 80\% & 80\% & 100\%  & 40\% & 60\% & 60\% & 20\% & 20\% & 40\% \\
			5 & 60\% & 100\% & 80\%  & 60\% & 100\% & 40\% & 40\% & 20\% & 0\% \\
			\bottomrule
		\end{tabular}
	\end{table}
	
	\subsection{Ablation: expanded-head baseline (no soft labels)}
	\label{sec:ablation-expanded-head}
	
	We verify that improvements are not simply due to increasing the output dimension. For this ablation we use FedAvg as the base optimizer and compare:
	\begin{enumerate}
		\item \textbf{FedAvg (baseline)}
		\item \textbf{FedAvg + expanded-head (hard):} final head expanded to $M\times C$ but trained with regular hard client-fixed joint labels.
		\item \textbf{FedAvg + JDFL (graded)} 
		\item \textbf{FedAvg + JDFL (random)}
	\end{enumerate}
	
	\textbf{Setup.} We run this ablation on PACS (setting B) and Office--Caltech using the same hyperparameters described in Section~\ref{sec:hyper}. Figure~\ref{fig:fedavg_comparison} presents the results. The comparison shows that merely expanding the classifier head and applying standard hard supervision does not account for the observed improvements. 
	Each reported entry is averaged over the same 3 random seeds used in the main experiments. 
	
	\subsection{Sensitivity to the number of pseudo-domains}
	\label{sec:sensitivity_M}
	
\begin{figure}[!ht]
		\centering
		\begin{subfigure}[t]{0.6\textwidth}
			\centering
			\begin{tikzpicture}
				\begin{axis}[
					ybar,
					bar width=7pt,
					width=\linewidth,
					height=6cm,
					ymin=60, ymax=80,
					ylabel={Class-Only Accuracy (\%)},
					symbolic x coords={
						{FedAvg\\(baseline)},
						{FedAvg + head expansion (hard)},
						{FedAvg + JDFL (graded)},
						{FedAvg + JDFL (random)}
					},
					xtick=data,
					enlarge x limits=0.12,
					xticklabel shift=3pt,
					xticklabel style={
						rotate=45,
						anchor=east,
						font=\scriptsize,
						text width=1.8cm,
						align=right
					},
					axis x line*=bottom,
					axis y line*=left,
					ymajorgrids=true,
					grid style=dashed,
					legend style={
						at={(0.5,1.1)},
						anchor=south,
						legend columns=2,
						font=\scriptsize
					},
					font=\scriptsize,
					]
					\addplot coordinates {
						({FedAvg\\(baseline)}, 73.10)
						({FedAvg + head expansion (hard)}, 71.33)
						({FedAvg + JDFL (graded)}, 76.51)
						({FedAvg + JDFL (random)}, 76.97)
					};
					\addplot coordinates {
						({FedAvg\\(baseline)}, 67.90)
						({FedAvg + head expansion (hard)}, 67.80)
						({FedAvg + JDFL (graded)}, 69.87)
						({FedAvg + JDFL (random)}, 70.53)
					};
					\legend{Office-Caltech, PACS (B)}
				\end{axis}
			\end{tikzpicture}
			\subcaption{Test accuracy in the ablation study. Expanding the classifier head alone with hard labels does not improve performance; combining the expansion with graded or random labeling yields clear gains.}
			\label{fig:fedavg_comparison}
		\end{subfigure}
		\hspace{0.02\textwidth}
		\begin{subfigure}[t]{0.6\textwidth}
			\centering
			\begin{tikzpicture}
				\begin{axis}[
					width=\linewidth,
					height=6cm,
					xlabel={Number of Pseudo-Domains (M)},
					ylabel={Global Test Accuracy (\%)},
					xmin=2, xmax=7,
					ymin=66, ymax=72,
					xtick={2,3,4,5,6,7},
					xticklabel style={font=\scriptsize},
					yticklabel style={font=\scriptsize},
					ymajorgrids=true,
					grid style=dashed,
					legend style={at={(0.5,1.1)}, anchor=south, legend columns=3, font=\scriptsize},
					every axis/.append style={font=\scriptsize},
					]
					\addplot[
					color=green!60!black,
					mark=o,
					thick,
					mark options={solid}
					] coordinates {
						(2,70.27) (3,70.57) (4,70.13) (5,70.40) (6,70.63) (7,70.63)
					};
					\addplot[
					color=red,
					mark=square*,
					thick,
					mark options={solid, fill=red}
					] coordinates {
						(2,69.40) (3,69.63) (4,69.97) (5,69.97) (6,70.33) (7,70.27)
					};
					\addplot[
					color=blue,
					dashed,
					thick
					] coordinates {(2,67.90) (7,67.90)};
					\legend{Random, Graded, Baseline (M=1)}
				\end{axis}
			\end{tikzpicture}
			\subcaption{Global test accuracy (on PACS setting B) as a function of the number of pseudo-domains $M \in \{2,\dots,7\}$.}
			\label{fig:pseudo_domains_accuracy}
		\end{subfigure}
		\caption{Ablation study and sensitivity analysis. (a) Comparison of FedAvg variants, showing that only the proposed JDFL schemes improve accuracy. (b) Performance robustness over different numbers of pseudo-domains $M$.}
		\label{fig:combined_ablation_sensitivity}
	\end{figure}
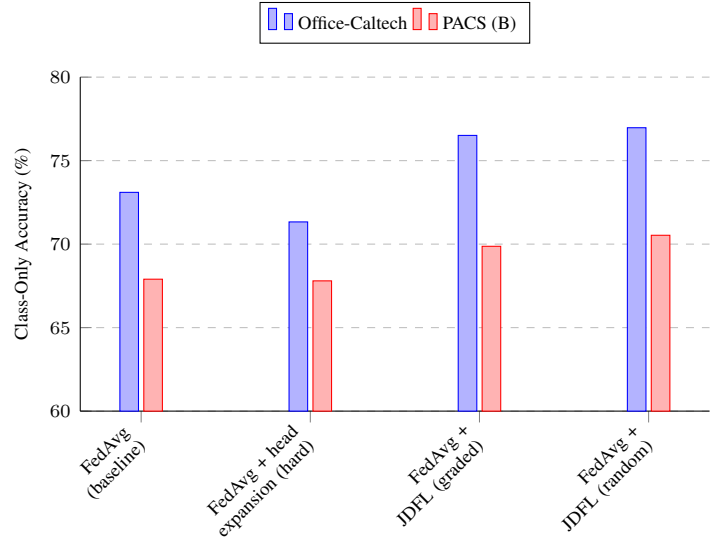
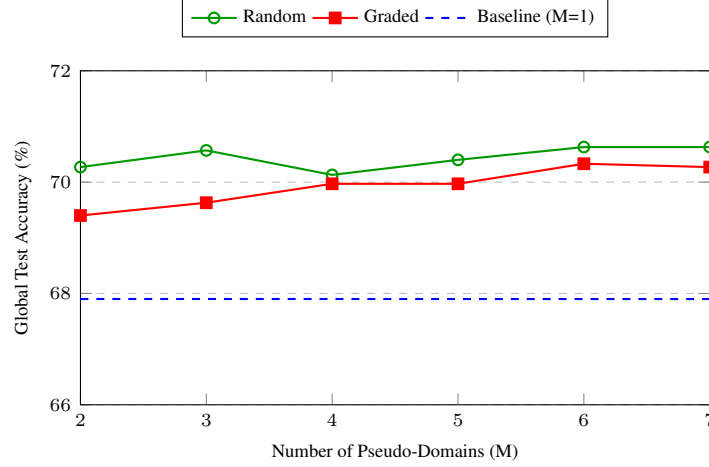

	To study how sensitive JDFL is to the choice of $M$, we sweep $M \in \{2,\dots,7\}$
	while the true number of domains in the data is 4 and report global test accuracy
	in Fig.~\ref{fig:pseudo_domains_accuracy}. The results show that performance is
	stable across a wide range of $M$: both randomized and graded supervision
	consistently outperform the $M=1$ baseline, and accuracy varies only
	marginally when $M$ deviates from the true domain count. This
	indicates that JDFL is robust to modest clustering errors.
	
	The absence of a sharp performance peak at the true domain count ($M=4$) highlights the built-in robustness of both supervision schemes. For Graded-JDFL, when $M$ exceeds the true number of domains, the resulting sub-domains naturally exhibit high centroid similarity. Consequently, the soft-target formulation dynamically shares substantial information across these artificially split clusters, implicitly re-merging them during optimization. For Randomized-JDFL, over-estimating $M$ simply increases the model's expanded capacity ($M \times C$). Rather than overfitting to these fragmented clusters, the randomized domain selection acts as a strong regularizer that prevents performance degradation, maintaining stable accuracy even with excess parameters.
	
	Despite this robustness, the Domain Discovery phase remains fundamentally necessary. For Graded-JDFL, computing domain centroids ($\mu_d$) in the feature space is a strict mathematical prerequisite, as the distances between these centroids dictate the inter-domain knowledge transfer. For Randomized-JDFL, while it does not rely on these topological distances, clustering provides a principled, data-driven method to establish baseline client assignments ($z_i$) and scale the network's capacity. Instead of guessing an arbitrary expansion size, the clustering phase guarantees that the parameter increase is grounded in the actual statistical diversity of the federated network, ensuring we neither bottleneck the model nor inflate parameters needlessly.
	
	\section{Conclusion and future directions}
	
	We presented \emph{JDFL}, a lightweight,
	optimizer-agnostic extension for handling feature skew in federated learning.
	JDFL uncovers client pseudo-domains via a short discovery phase and expands
	the classifier head to \(M\times C\) joint logits. We study two supervision
	schemes on this head - \emph{graded invariance}, which introduces the notion of
	graded domain invariance (domain-aware sharing) into the federated learning setup; and
	\emph{randomized joint-label assignment}, which leverages the non-linear,
	piecewise structure of the expanded final layer as a training-time regularizer
	by stochastically perturbing supervision - and show that both improve
	class-only generalization under domain heterogeneity, with different trade-offs
	between specialization and regularization.
	
	Future work should make the similarity-based soft-labeling adaptive to achieve a special goal for example domain invariance. Rather
	than using a fixed centroid-derived distribution, one could learn a dynamic
	domain posterior for the expanded head so that transfer weights evolve with
	training. Future work on the systems side, could relax the requirement that all clients must
	participate in an initial clustering pass. Practical alternatives include
	clustering a sampled subset of representative clients; clustering on compact
	``sketches'' of updates; or performing incremental clustering that
	updates centroids as clients join or reappear to be able to use JDFL in realistic cross-device sceinarios more effectively (Although we can use the initial subset of clients update vectors in clustering in cross-device cases but it definitely could become more efficient).
	Finally With this expanded head, another idea is to add
	regularization terms in order to encourage some desired behaviour during the training
	of the expanded model to learn high-quality representations with higher generalizaton and therefore achieve better results. We think investigating this could lead to great achievements.
	
	We hope our findings help inspire subsequent efforts in the field of heterogeneous federated learning.
	
	\clearpage
	
	\bibliographystyle{unsrt}  
	\bibliography{references}  
	
	\clearpage
	
	
	\appendix
		
	\input{appendix}


\end{document}

%% file: appendix.tex
\section{On the nonlinearity induced by expanding the output layer}
\label{app:A}
In standard linear classifiers each semantic class $c$ is represented by a single affine score
\[
z_c(h) = w_c^\top h + b_c,
\]
where \(h\in\mathbb{R}^p\) denotes the feature vector produced by the network backbone,
and \(w_c\in\mathbb{R}^p, b_c\in\mathbb{R}\) are the class-specific parameters. The map
\(h\mapsto z_c(h)\) is affine (linear plus shift), and the two-class decision boundary
between classes \(c\) and \(c'\) is the hyperplane
\(\{h : z_c(h)=z_{c'}(h)\}\). Thus, with a single output per class the classifier implements
a \emph{linear} decision rule in feature space.

When the final layer is expanded to have \(M\) units per class, the construction changes
fundamentally. For each class \(c\) we now have \(M\) affine outputs
\[
\ell_{c,1}(h)=w_{c,1}^\top h + b_{c,1},\quad
\ldots,\quad
\ell_{c,M}(h)=w_{c,M}^\top h + b_{c,M}.
\]
We then define the class score by taking the maximum over those \(M\) outputs:
\[
s_c(h)=\max_{j=1,\dots,M}\ell_{c,j}(h).
\]
Prediction is made by selecting the class with largest score,
\(\hat c=\arg\max_{c\in\{1,\dots,C\}} s_c(h)\).

The crucial observation is that each \(s_c(h)\) is a \emph{max of affine} functions.
Consequently:
\begin{itemize}
	\item \(s_c(h)\) is a convex, \emph{piecewise-linear} function of \(h\): the feature space
	is partitioned into polyhedral regions on each of which a single affine function
	\(\ell_{c,j}\) is active and equals \(s_c(h)\).
	\item The decision boundary between two classes \(c\) and \(c'\) is
	\[
	\{h : s_c(h)=s_{c'}(h)\}
	\subseteq \bigcup_{j=1}^M\bigcup_{k=1}^M \{\,h : \ell_{c,j}(h)=\ell_{c',k}(h)\,\},
	\]
	and it consists of those linear pieces of the hyperplanes \(\ell_{c,j}(h)=\ell_{c',k}(h)\)
	where \(\ell_{c,j}\) and \(\ell_{c',k}\) are the respective active maximisers for classes
	\(c\) and \(c'\). This boundary is therefore not a single hyperplane but a
	\emph{piecewise-linear} surface (a finite union of linear facets, at most
	\(M^2\) per class pair).
\end{itemize}

Because the \(\max\) operator is nonlinear, the classifier obtained by combining
the expanded linear layer with the max-and-argmax prediction rule is itself
\emph{piecewise-linear} and therefore \emph{nonlinear} in the usual sense
(although the parameter-to-logit map comprising the affine outputs remains linear).
A convenient way to think of this construction is as a \emph{maxout-like} head
\cite{goodfellow2013maxout}: unlike the original maxout which is typically applied as a per-unit activation in a hidden layer, our construction applies maxout \emph{per class} — i.e., we take the max of $M$ affine scorers for each semantic class. (multiple affine scorers per class are available, and the max operator selects the most relevant scorer for each input.)

\paragraph{Training implications.}
It is important to distinguish training from inference. During training, the loss is applied directly to the full set of 
$M \times C$ logits, without any max operation inside the computational graph. Gradients therefore propagate to all logits according
to the chosen probabilistic supervision (graded or randomized), and no nondifferentiability arises in the optimization
process. Nevertheless, the inference rule induces a piecewise-linear partition of feature space, and the expanded outputs
can implicitly specialize to different appearance modes. The supervision schemes encourage these multiple affine scorers
to remain compatible and to share responsibility for modeling a class, preventing any single scorer from dominating.

\paragraph{A simple numeric (1-D) example.}
Let \(h\in\mathbb{R}\) be a scalar feature and consider a single class with \(M=2\)
affine outputs
\[
\ell_1(h) = h,\qquad \ell_2(h) = -h + 2.
\]
The class score is
\[
s(h)=\max(\ell_1(h),\ell_2(h))=\max(h,-h+2).
\]
Evaluate:
\begin{itemize}
	\item If \(h<1\), then \(-h+2>h\), so \(s(h)=-h+2\) (an affine function with slope \(-1\)).
	\item If \(h>1\), then \(h>-h+2\), so \(s(h)=h\) (an affine function with slope \(+1\)).
	\item At \(h=1\) the two affines are equal; the function switches active branch.
\end{itemize}
Hence $s(h)$ is a ``V''-shaped piecewise-linear function, which cannot be
represented by a single affine function and therefore illustrates the
increased expressivity obtained by multiple outputs per class.

\paragraph{Summary and Implications.}
Although the final weight layer is implemented as a linear map that outputs
\(M\times C\) logits, the prediction rule (taking the maximum over the \(M\)
units per class and then the class-wise argmax) makes the overall classifier a
\emph{piecewise-linear} (max-of-affine) function of the features. Consequently,
the classifier is nonlinear in its mapping from features to discrete labels,
and this increased piecewise-linear expressivity is a principal mechanism by
which multiple units per class can model multimodal appearance and interact
with the graded and randomized supervision schemes introduced in this work.

\begin{table*}[t]
	\centering
	\caption{Detailed architecture of the model used for experiments. This is a typical CNN architecture that performs well on the bench datasets we use for this paper. }
	\label{tab:architecture-my-model}
	\footnotesize
	\setlength{\tabcolsep}{4pt}
	\begin{tabular}{@{}llcl@{}}
		\toprule
		\textbf{Block} & \textbf{Layer} & \textbf{Output Shape} & \textbf{Details} \\
		\midrule
		\multirow{5}{*}{Block 1}
		& Conv2d(3, 32, 3, padding=1)   & (32, $H$, $W$)     & kernel=3$\times$3, pad=1 \\
		& GroupNorm(8, 32) + ReLU        & (32, $H$, $W$)     & 8 groups \\
		& Conv2d(32, 32, 3, padding=1)  & (32, $H$, $W$)     & kernel=3$\times$3, pad=1 \\
		& GroupNorm(8, 32) + ReLU        & (32, $H$, $W$)     & 8 groups \\
		& MaxPool2d(2, 2)                & (32, $H$/2, $W$/2) & stride=2 \\
		\midrule
		\multirow{5}{*}{Block 2}
		& Conv2d(32, 64, 3, padding=1)  & (64, $H$/2, $W$/2) & kernel=3$\times$3, pad=1 \\
		& GroupNorm(16, 64) + ReLU       & (64, $H$/2, $W$/2) & 16 groups \\
		& Conv2d(64, 64, 3, padding=1)  & (64, $H$/2, $W$/2) & kernel=3$\times$3, pad=1 \\
		& GroupNorm(16, 64) + ReLU       & (64, $H$/2, $W$/2) & 16 groups \\
		& MaxPool2d(2, 2)                & (64, $H$/4, $W$/4) & stride=2 \\
		\midrule
		\multirow{5}{*}{Block 3}
		& Conv2d(64, 128, 3, padding=1) & (128, $H$/4, $W$/4) & kernel=3$\times$3, pad=1 \\
		& GroupNorm(16, 128) + ReLU      & (128, $H$/4, $W$/4) & 16 groups \\
		& Conv2d(128, 128, 3, padding=1)& (128, $H$/4, $W$/4) & kernel=3$\times$3, pad=1 \\
		& GroupNorm(16, 128) + ReLU      & (128, $H$/4, $W$/4) & 16 groups \\
		& MaxPool2d(2, 2)                & (128, $H$/8, $W$/8)& stride=2 \\
		\midrule
		\multirow{3}{*}{FC Layers}
		& Linear(128\,$\times$\,$HW$/64, 256) & (256)   & flatten + fc1 \\
		& Linear(256, 128)                & (128)   & fc2 \\
		& Linear(128, $C$)                & ($C$)   & fc3 \\
		\bottomrule
	\end{tabular}
\end{table*}

\section{Model Architecture}
\label{app:B}
The architecture of the base-line model used for all the experiments is available in Table~\ref{tab:architecture-my-model}. The input height and width
We used is (114, 114).


\section{Hyperparameters and training history plots}
\label{app:C}
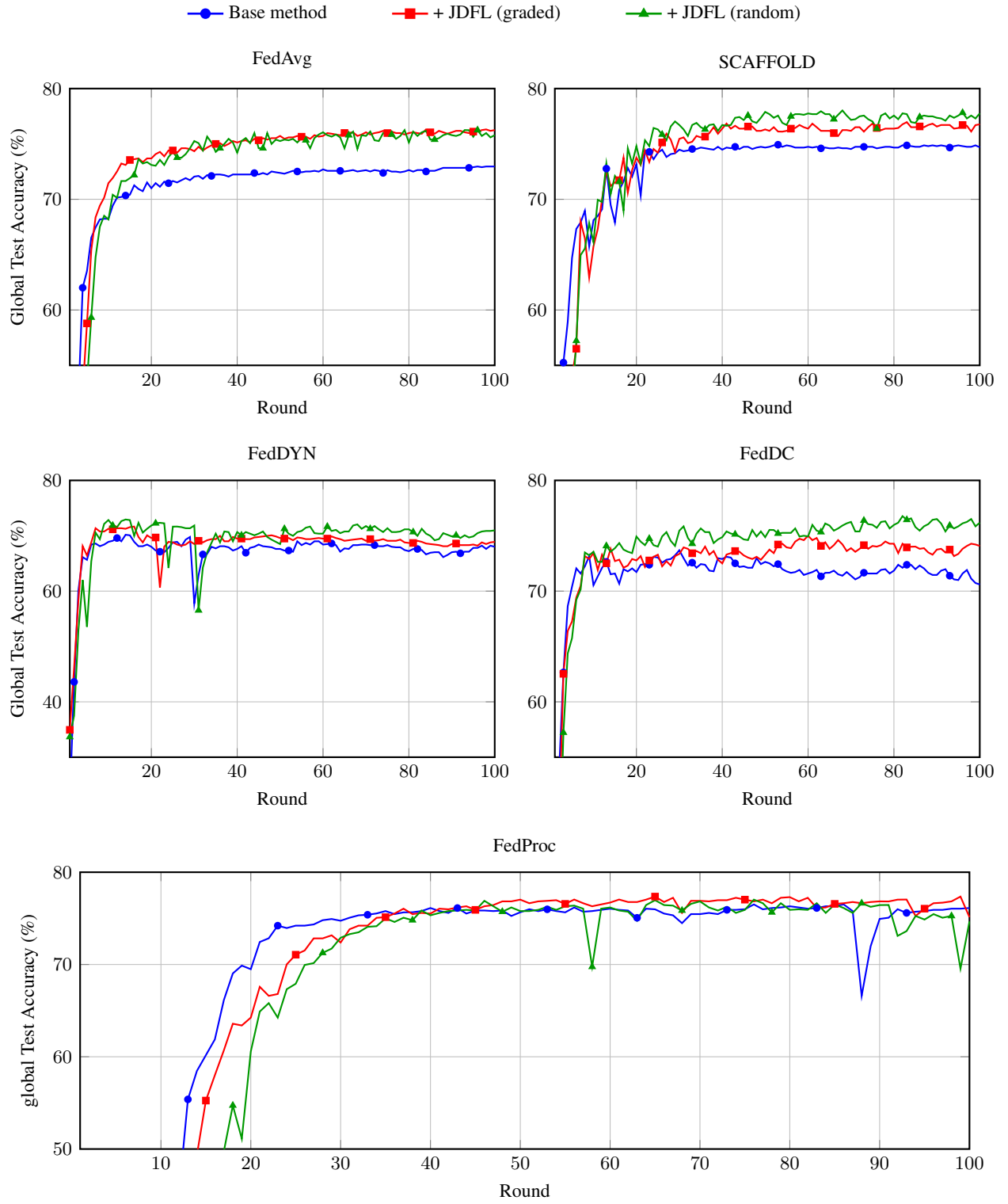
\begin{figure*}[t]
	\centering
	\setlength{\tabcolsep}{1pt}
	
	\begin{tikzpicture}
		\begin{axis}[
			hide axis,
			xmin=0,xmax=1,ymin=0,ymax=1,
			legend columns=3,
			legend style={
				at={(0.5,0.9)},
				anchor=south,
				draw=none,
				font=\small,
				/tikz/every even column/.append style={column sep=1.0cm}
			}
			]
			\addlegendimage{color=blue,line width=1pt, mark=*}\addlegendentry{Base method}
			\addlegendimage{color=red,line width=1pt, mark=square*}\addlegendentry{+ JDFL (graded)}
			\addlegendimage{color=green!60!black,line width=1pt, mark=triangle*}\addlegendentry{+ JDFL (random)}
		\end{axis}
	\end{tikzpicture}
	
	\vspace{2mm} 
	
	\begin{tabular}{cc}
		\begin{tikzpicture}
			\begin{axis}[
				width=0.43\textwidth,
				height=0.28\textwidth,
				xmin=1, xmax=100,
				ymin=55, ymax=80,    
				xlabel=Round,
				ylabel={Global Test Accuracy (\%)},
				title={FedAvg},
				grid=both,
				grid style={line width=.1pt, draw=gray!30},
				major grid style={line width=.2pt, draw=gray!60},
				mark repeat=10,
				mark size=1.4pt,
				thick,
				scale only axis,
				tick label style={font=\small},
				label style={font=\small},
				title style={font=\small}
				]
				\addplot[color=blue, mark=*] table[x index=0, y=FedAvg_base, col sep=comma] {officedata.csv};
				\addplot[color=red, mark=square*] table[x index=0, y=FedAvg_graded, col sep=comma] {officedata.csv};
				\addplot[color=green!60!black, mark=triangle*] table[x index=0, y=FedAvg_random, col sep=comma] {officedata.csv};
			\end{axis}
		\end{tikzpicture}
		&
		\begin{tikzpicture}
			\begin{axis}[
				width=0.43\textwidth,
				height=0.28\textwidth,
				xmin=1, xmax=100,
				ymin=55, ymax=80,
				xlabel=Round,
				ylabel={}, 
				title={SCAFFOLD},
				grid=both,
				mark repeat=10,
				mark size=1.4pt,
				thick,
				scale only axis,
				tick label style={font=\small},
				label style={font=\small},
				title style={font=\small}
				]
				\addplot[color=blue, mark=*] table[x index=0, y=SCAFFOLD_base, col sep=comma] {officedata.csv};
				\addplot[color=red, mark=square*] table[x index=0, y=SCAFFOLD_graded, col sep=comma] {officedata.csv};
				\addplot[color=green!60!black, mark=triangle*] table[x index=0, y=SCAFFOLD_random, col sep=comma] {officedata.csv};
			\end{axis}
		\end{tikzpicture}
		\\[6pt]
		
		\begin{tikzpicture}
			\begin{axis}[
				width=0.43\textwidth,
				height=0.28\textwidth,
				xmin=1, xmax=100,
				ymin=30, ymax=80,
				xlabel=Round,
				ylabel={Global Test Accuracy (\%)},
				title={FedDYN},
				grid=both,
				mark repeat=10,
				mark size=1.4pt,
				thick,
				scale only axis,
				tick label style={font=\small},
				label style={font=\small},
				title style={font=\small}
				]
				\addplot[color=blue, mark=*] table[x index=0, y=FedDYN_base, col sep=comma] {officedata.csv};
				\addplot[color=red, mark=square*] table[x index=0, y=FedDYN_graded, col sep=comma] {officedata.csv};
				\addplot[color=green!60!black, mark=triangle*] table[x index=0, y=FedDYN_random, col sep=comma] {officedata.csv};
			\end{axis}
		\end{tikzpicture}
		&
		\begin{tikzpicture}
			\begin{axis}[
				width=0.43\textwidth,
				height=0.28\textwidth,
				xmin=1, xmax=100,
				ymin=55, ymax=80,
				xlabel=Round,
				ylabel={}, 
				title={FedDC},
				grid=both,
				mark repeat=10,
				mark size=1.4pt,
				thick,
				scale only axis,
				tick label style={font=\small},
				label style={font=\small},
				title style={font=\small}
				]
				\addplot[color=blue, mark=*] table[x index=0, y=FedDC_base, col sep=comma] {officedata.csv};
				\addplot[color=red, mark=square*] table[x index=0, y=FedDC_graded, col sep=comma] {officedata.csv};
				\addplot[color=green!60!black, mark=triangle*] table[x index=0, y=FedDC_random, col sep=comma] {officedata.csv};
			\end{axis}
		\end{tikzpicture}
		\\[6pt]
		
		\multicolumn{2}{c}{
			\begin{tikzpicture}
				\begin{axis}[
					width=0.9\textwidth,     
					height=0.28\textwidth,
					xmin=1, xmax=100,
					ymin=50, ymax=80,
					xlabel=Round,
					ylabel={global Test Accuracy (\%)},
					title={FedProc},
					grid=both,
					mark repeat=10,
					mark size=1.4pt,
					thick,
					scale only axis,
					tick label style={font=\small},
					label style={font=\small},
					title style={font=\small}
					]
					\addplot[color=blue, mark=*] table[x index=0, y=FedProc_base, col sep=comma] {officedata.csv};
					\addplot[color=red, mark=square*] table[x index=0, y=FedProc_graded, col sep=comma] {officedata.csv};
					\addplot[color=green!60!black, mark=triangle*] table[x index=0, y=FedProc_random, col sep=comma] {officedata.csv};
				\end{axis}
			\end{tikzpicture}
		}
	\end{tabular}
	
	\caption{Test accuracy history — Experiment on Office-Caltech. Each plot shows the base method and two JDFL variants (graded, random).}
	\label{fig:compact-panels_office}
\end{figure*}

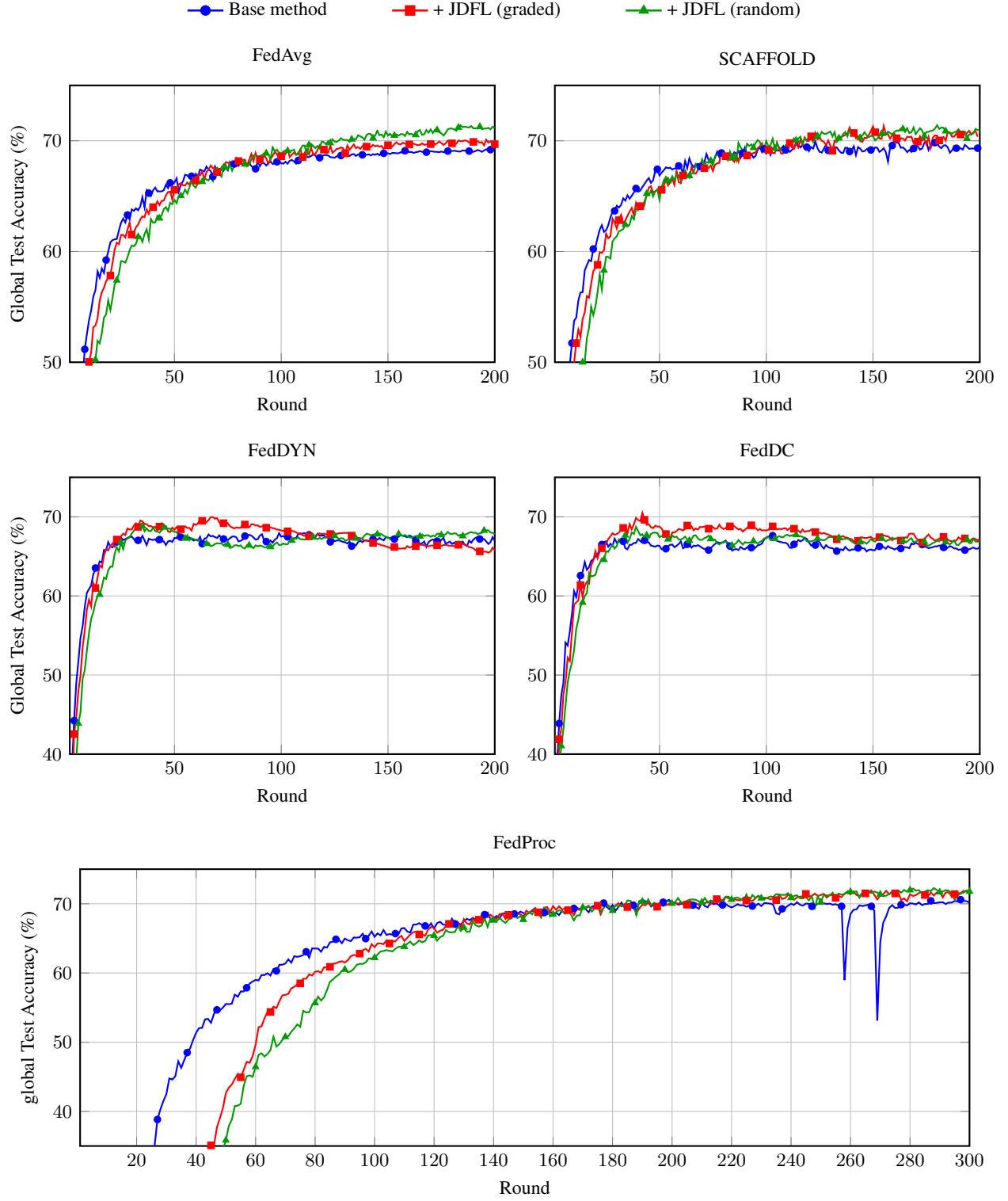
\begin{figure*}[t]
	\centering
	\setlength{\tabcolsep}{1pt}
	
	\begin{tikzpicture}
		\begin{axis}[
			hide axis,
			xmin=0,xmax=1,ymin=0,ymax=1,
			legend columns=3,
			legend style={
				at={(0.5,0.9)},
				anchor=south,
				draw=none,
				font=\small,
				/tikz/every even column/.append style={column sep=1.0cm}
			}
			]
			\addlegendimage{color=blue,line width=1pt, mark=*}\addlegendentry{Base method}
			\addlegendimage{color=red,line width=1pt, mark=square*}\addlegendentry{+ JDFL (graded)}
			\addlegendimage{color=green!60!black,line width=1pt, mark=triangle*}\addlegendentry{+ JDFL (random)}
		\end{axis}
	\end{tikzpicture}
	
	\vspace{2mm} 
	
	\begin{tabular}{cc}
		\begin{tikzpicture}
			\begin{axis}[
				width=0.43\textwidth,
				height=0.28\textwidth,
				xmin=1, xmax=200,
				ymin=50, ymax=75,    
				xlabel=Round,
				ylabel={Global Test Accuracy (\%)},
				title={FedAvg},
				grid=both,
				grid style={line width=.1pt, draw=gray!30},
				major grid style={line width=.2pt, draw=gray!60},
				mark repeat=10,
				mark size=1.4pt,
				thick,
				scale only axis,
				tick label style={font=\small},
				label style={font=\small},
				title style={font=\small}
				]
				\addplot[color=blue, mark=*] table[x index=0, y=FedAvg_base, col sep=comma] {pacsadata.csv};
				\addplot[color=red, mark=square*] table[x index=0, y=FedAvg_graded, col sep=comma] {pacsadata.csv};
				\addplot[color=green!60!black, mark=triangle*] table[x index=0, y=FedAvg_random, col sep=comma] {pacsadata.csv};
			\end{axis}
		\end{tikzpicture}
		&
		\begin{tikzpicture}
			\begin{axis}[
				width=0.43\textwidth,
				height=0.28\textwidth,
				xmin=1, xmax=200,
				ymin=50, ymax=75,
				xlabel=Round,
				ylabel={}, 
				title={SCAFFOLD},
				grid=both,
				mark repeat=10,
				mark size=1.4pt,
				thick,
				scale only axis,
				tick label style={font=\small},
				label style={font=\small},
				title style={font=\small}
				]
				\addplot[color=blue, mark=*] table[x index=0, y=SCAFFOLD_base, col sep=comma] {pacsadata.csv};
				\addplot[color=red, mark=square*] table[x index=0, y=SCAFFOLD_graded, col sep=comma] {pacsadata.csv};
				\addplot[color=green!60!black, mark=triangle*] table[x index=0, y=SCAFFOLD_random, col sep=comma] {pacsadata.csv};
			\end{axis}
		\end{tikzpicture}
		\\[6pt]
		
		\begin{tikzpicture}
			\begin{axis}[
				width=0.43\textwidth,
				height=0.28\textwidth,
				xmin=1, xmax=200,
				ymin=40, ymax=75,
				xlabel=Round,
				ylabel={Global Test Accuracy (\%)},
				title={FedDYN},
				grid=both,
				mark repeat=10,
				mark size=1.4pt,
				thick,
				scale only axis,
				tick label style={font=\small},
				label style={font=\small},
				title style={font=\small}
				]
				\addplot[color=blue, mark=*] table[x index=0, y=FedDYN_base, col sep=comma] {pacsadata.csv};
				\addplot[color=red, mark=square*] table[x index=0, y=FedDYN_graded, col sep=comma] {pacsadata.csv};
				\addplot[color=green!60!black, mark=triangle*] table[x index=0, y=FedDYN_random, col sep=comma] {pacsadata.csv};
			\end{axis}
		\end{tikzpicture}
		&
		\begin{tikzpicture}
			\begin{axis}[
				width=0.43\textwidth,
				height=0.28\textwidth,
				xmin=1, xmax=200,
				ymin=40, ymax=75,
				xlabel=Round,
				ylabel={}, 
				title={FedDC},
				grid=both,
				mark repeat=10,
				mark size=1.4pt,
				thick,
				scale only axis,
				tick label style={font=\small},
				label style={font=\small},
				title style={font=\small}
				]
				\addplot[color=blue, mark=*] table[x index=0, y=FedDC_base, col sep=comma] {pacsadata.csv};
				\addplot[color=red, mark=square*] table[x index=0, y=FedDC_graded, col sep=comma] {pacsadata.csv};
				\addplot[color=green!60!black, mark=triangle*] table[x index=0, y=FedDC_random, col sep=comma] {pacsadata.csv};
			\end{axis}
		\end{tikzpicture}
		\\[6pt]
		
		\multicolumn{2}{c}{
			\begin{tikzpicture}
				\begin{axis}[
					width=0.9\textwidth,     
					height=0.28\textwidth,
					xmin=1, xmax=300,
					ymin=35, ymax=75,
					xlabel=Round,
					ylabel={global Test Accuracy (\%)},
					title={FedProc},
					grid=both,
					mark repeat=10,
					mark size=1.4pt,
					thick,
					scale only axis,
					tick label style={font=\small},
					label style={font=\small},
					title style={font=\small}
					]
					\addplot[color=blue, mark=*] table[x index=0, y=FedProc_base, col sep=comma] {pacsadata.csv};
					\addplot[color=red, mark=square*] table[x index=0, y=FedProc_graded, col sep=comma] {pacsadata.csv};
					\addplot[color=green!60!black, mark=triangle*] table[x index=0, y=FedProc_random, col sep=comma] {pacsadata.csv};
				\end{axis}
			\end{tikzpicture}
		}
	\end{tabular}
	
	\caption{Test accuracy history — Experiment on PACS(A). Each plot shows the base method and two JDFL variants (graded, random).}
	\label{fig:compact-panels_PACSA}
\end{figure*}

\begin{figure*}[t]
	\centering
	\setlength{\tabcolsep}{0.5pt}
	
	\begin{tikzpicture}
		\begin{axis}[
			hide axis,
			xmin=0,xmax=1,ymin=0,ymax=1,
			legend columns=3,
			legend style={
				at={(0.5,0.9)},
				anchor=south,
				draw=none,
				font=\small,
				/tikz/every even column/.append style={column sep=1.0cm}
			}
			]
			\addlegendimage{color=blue,line width=1pt, mark=*}\addlegendentry{Base method}
			\addlegendimage{color=red,line width=1pt, mark=square*}\addlegendentry{+ JDFL (graded)}
			\addlegendimage{color=green!60!black,line width=1pt, mark=triangle*}\addlegendentry{+ JDFL (random)}
		\end{axis}
	\end{tikzpicture}
	
	\vspace{2mm} 
	
	\begin{tabular}{cc}
		\begin{tikzpicture}
			\begin{axis}[
				width=0.43\textwidth,
				height=0.28\textwidth,
				xmin=1, xmax=150,
				ymin=55, ymax=75,    
				xlabel=Round,
				ylabel={Global Test Accuracy (\%)},
				title={FedAvg},
				grid=both,
				grid style={line width=.1pt, draw=gray!30},
				major grid style={line width=.2pt, draw=gray!60},
				mark repeat=10,
				mark size=1.4pt,
				thick,
				scale only axis,
				tick label style={font=\small},
				label style={font=\small},
				title style={font=\small}
				]
				\addplot[color=blue, mark=*] table[x index=0, y=FedAvg_base, col sep=comma] {pacsbdata.csv};
				\addplot[color=red, mark=square*] table[x index=0, y=FedAvg_graded, col sep=comma] {pacsbdata.csv};
				\addplot[color=green!60!black, mark=triangle*] table[x index=0, y=FedAvg_random, col sep=comma] {pacsbdata.csv};
			\end{axis}
		\end{tikzpicture}
		&
		\begin{tikzpicture}
			\begin{axis}[
				width=0.43\textwidth,
				height=0.28\textwidth,
				xmin=1, xmax=150,
				ymin=55, ymax=75,
				xlabel=Round,
				ylabel={}, 
				title={SCAFFOLD},
				grid=both,
				mark repeat=10,
				mark size=1.4pt,
				thick,
				scale only axis,
				tick label style={font=\small},
				label style={font=\small},
				title style={font=\small}
				]
				\addplot[color=blue, mark=*] table[x index=0, y=SCAFFOLD_base, col sep=comma] {pacsbdata.csv};
				\addplot[color=red, mark=square*] table[x index=0, y=SCAFFOLD_graded, col sep=comma] {pacsbdata.csv};
				\addplot[color=green!60!black, mark=triangle*] table[x index=0, y=SCAFFOLD_random, col sep=comma] {pacsbdata.csv};
			\end{axis}
		\end{tikzpicture}
		\\[6pt]
		
		\begin{tikzpicture}
			\begin{axis}[
				width=0.43\textwidth,
				height=0.28\textwidth,
				xmin=1, xmax=150,
				ymin=55, ymax=70,
				xlabel=Round,
				ylabel={Global Test Accuracy (\%)},
				title={FedDYN},
				grid=both,
				mark repeat=10,
				mark size=1.4pt,
				thick,
				scale only axis,
				tick label style={font=\small},
				label style={font=\small},
				title style={font=\small}
				]
				\addplot[color=blue, mark=*] table[x index=0, y=FedDYN_base, col sep=comma] {pacsbdata.csv};
				\addplot[color=red, mark=square*] table[x index=0, y=FedDYN_graded, col sep=comma] {pacsbdata.csv};
				\addplot[color=green!60!black, mark=triangle*] table[x index=0, y=FedDYN_random, col sep=comma] {pacsbdata.csv};
			\end{axis}
		\end{tikzpicture}
		&
		\begin{tikzpicture}
			\begin{axis}[
				width=0.43\textwidth,
				height=0.28\textwidth,
				xmin=1, xmax=150,
				ymin=55, ymax=75,
				xlabel=Round,
				ylabel={}, 
				title={FedDC},
				grid=both,
				mark repeat=10,
				mark size=1.4pt,
				thick,
				scale only axis,
				tick label style={font=\small},
				label style={font=\small},
				title style={font=\small}
				]
				\addplot[color=blue, mark=*] table[x index=0, y=FedDC_base, col sep=comma] {pacsbdata.csv};
				\addplot[color=red, mark=square*] table[x index=0, y=FedDC_graded, col sep=comma] {pacsbdata.csv};
				\addplot[color=green!60!black, mark=triangle*] table[x index=0, y=FedDC_random, col sep=comma] {pacsbdata.csv};
			\end{axis}
		\end{tikzpicture}
		\\[6pt]
		
		\multicolumn{2}{c}{
			\begin{tikzpicture}
				\begin{axis}[
					width=0.9\textwidth,     
					height=0.28\textwidth,
					xmin=1, xmax=150,
					ymin=50, ymax=75,
					xlabel=Round,
					ylabel={global Test Accuracy (\%)},
					title={FedProc},
					grid=both,
					mark repeat=10,
					mark size=1.4pt,
					thick,
					scale only axis,
					tick label style={font=\small},
					label style={font=\small},
					title style={font=\small}
					]
					\addplot[color=blue, mark=*] table[x index=0, y=FedProc_base, col sep=comma] {pacsbdata.csv};
					\addplot[color=red, mark=square*] table[x index=0, y=FedProc_graded, col sep=comma] {pacsbdata.csv};
					\addplot[color=green!60!black, mark=triangle*] table[x index=0, y=FedProc_random, col sep=comma] {pacsbdata.csv};
				\end{axis}
			\end{tikzpicture}
		}
	\end{tabular}
	
	\caption{Test accuracy history — Experiment on PACS (B). Each plot shows the base method and two JDFL variants (graded, random).}
	\label{fig:compact-panels_PACSB}
\end{figure*}
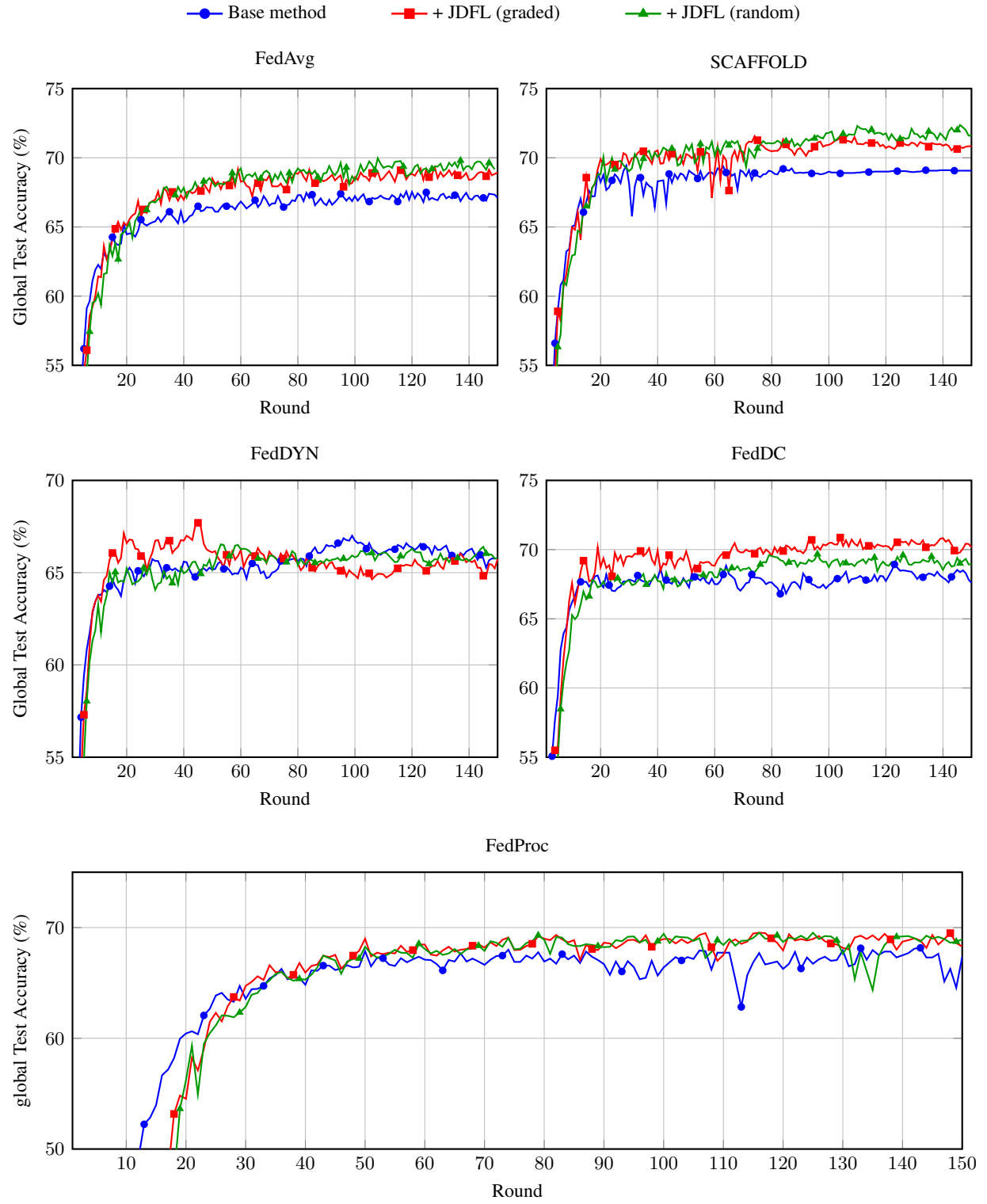

The training history for all the experiments is shown in Figures~\ref{fig:compact-panels_office}, ~\ref{fig:compact-panels_PACSA} and ~\ref{fig:compact-panels_PACSB} and the final results including the performance on each domain is reported in Tables~\ref{tab:office_e1},~\ref{tab:pacs_e1}, ~\ref{tab:pacs_e2} and ~\ref{tab:pacs_e3}.
In the experiments section we mentioned the optimizer and local learning rate for all experiments are SGD and 0.01. Also we talked about the hyperparameters related to JDFL extensions. Here we list the hyperparameters particular to each method that we used to achieve the training histories in Figures~\ref{fig:compact-panels_office}, ~\ref{fig:compact-panels_PACSA}, ~\ref{fig:compact-panels_PACSB} and ~\ref{fig:compact-panels_PACSC} below:

\begin{enumerate}
	\item SCAFFOLD for all the experiments: global learning rate = 1
	\item FedDYN for all experiments: regularization coefficient (alpha) = 0.01
	\item FedDC for PACS with setting A: regularization coefficient (alpha) = 0.01
	\item FedDC for Office-Caltech and PACS with setting B: regularization coefficient (alpha) = 0.1
\end{enumerate}

As we mentioned before we use the same hyperparameters for baseline methods and baseline methods combined with JDFL. The figures verify the fact that different baselines when combined with JDFL converge to higher accuracy in most cases and based on the tables achieve higher accuracy on test set in all cases.

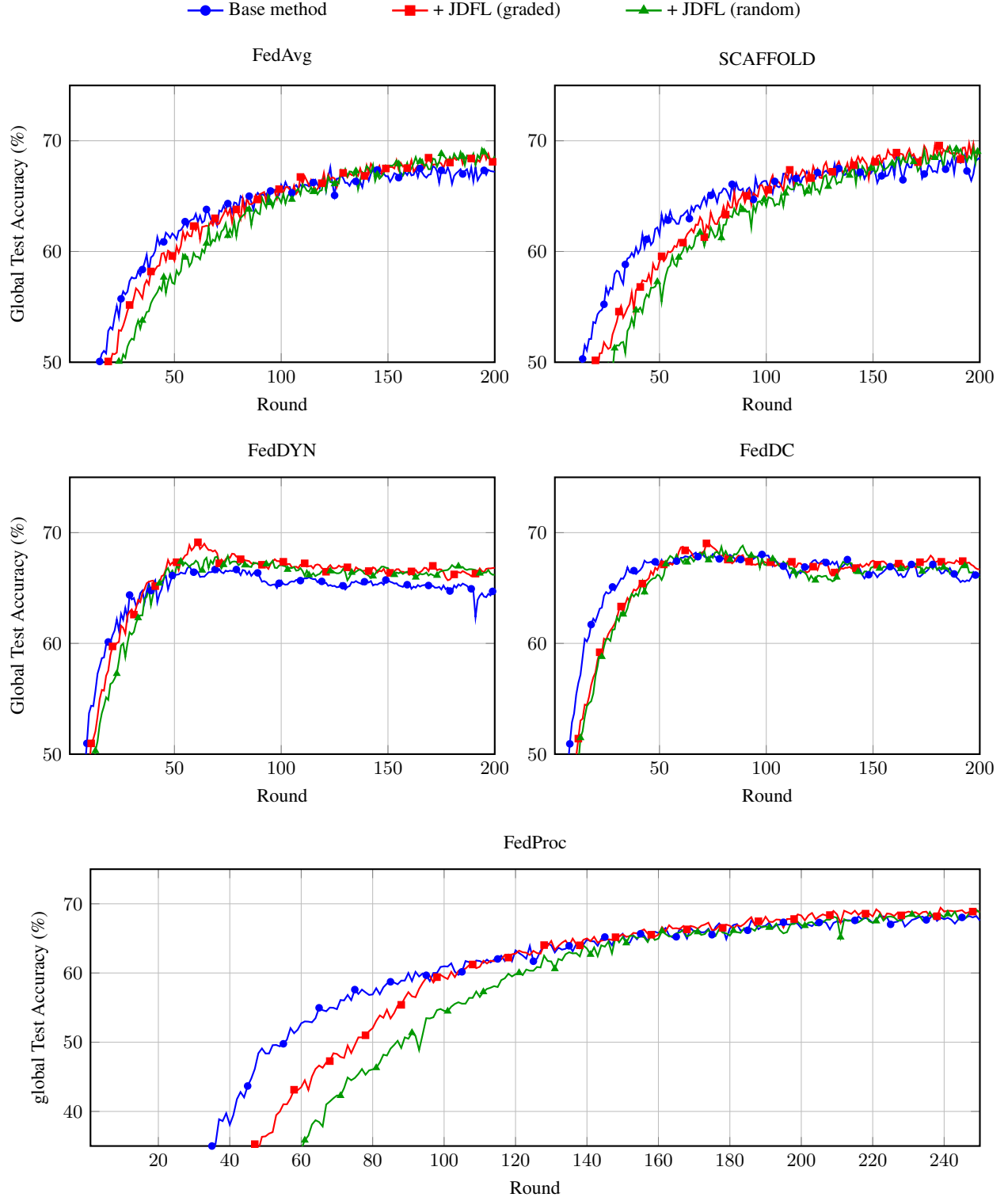
\begin{figure*}[t]
	\centering
	\setlength{\tabcolsep}{1pt}
	
	\begin{tikzpicture}
		\begin{axis}[
			hide axis,
			xmin=0,xmax=1,ymin=0,ymax=1,
			legend columns=3,
			legend style={
				at={(0.5,0.9)},
				anchor=south,
				draw=none,
				font=\small,
				/tikz/every even column/.append style={column sep=1.0cm}
			}
			]
			\addlegendimage{color=blue,line width=1pt, mark=*}\addlegendentry{Base method}
			\addlegendimage{color=red,line width=1pt, mark=square*}\addlegendentry{+ JDFL (graded)}
			\addlegendimage{color=green!60!black,line width=1pt, mark=triangle*}\addlegendentry{+ JDFL (random)}
		\end{axis}
	\end{tikzpicture}
	
	\vspace{2mm} 
	
	\begin{tabular}{cc}
		\begin{tikzpicture}
			\begin{axis}[
				width=0.43\textwidth,
				height=0.28\textwidth,
				xmin=1, xmax=200,
				ymin=50, ymax=75,    
				xlabel=Round,
				ylabel={Global Test Accuracy (\%)},
				title={FedAvg},
				grid=both,
				grid style={line width=.1pt, draw=gray!30},
				major grid style={line width=.2pt, draw=gray!60},
				mark repeat=10,
				mark size=1.4pt,
				thick,
				scale only axis,
				tick label style={font=\small},
				label style={font=\small},
				title style={font=\small}
				]
				\addplot[color=blue, mark=*] table[x index=0, y=FedAvg_base, col sep=comma] {pacsdatac.csv};
				\addplot[color=red, mark=square*] table[x index=0, y=FedAvg_graded, col sep=comma] {pacsdatac.csv};
				\addplot[color=green!60!black, mark=triangle*] table[x index=0, y=FedAvg_random, col sep=comma] {pacsdatac.csv};
			\end{axis}
		\end{tikzpicture}
		&
		\begin{tikzpicture}
			\begin{axis}[
				width=0.43\textwidth,
				height=0.28\textwidth,
				xmin=1, xmax=200,
				ymin=50, ymax=75,
				xlabel=Round,
				ylabel={}, 
				title={SCAFFOLD},
				grid=both,
				mark repeat=10,
				mark size=1.4pt,
				thick,
				scale only axis,
				tick label style={font=\small},
				label style={font=\small},
				title style={font=\small}
				]
				\addplot[color=blue, mark=*] table[x index=0, y=SCAFFOLD_base, col sep=comma] {pacsdatac.csv};
				\addplot[color=red, mark=square*] table[x index=0, y=SCAFFOLD_graded, col sep=comma] {pacsdatac.csv};
				\addplot[color=green!60!black, mark=triangle*] table[x index=0, y=SCAFFOLD_random, col sep=comma] {pacsdatac.csv};
			\end{axis}
		\end{tikzpicture}
		\\[6pt]
		
		\begin{tikzpicture}
			\begin{axis}[
				width=0.43\textwidth,
				height=0.28\textwidth,
				xmin=1, xmax=200,
				ymin=50, ymax=75,
				xlabel=Round,
				ylabel={Global Test Accuracy (\%)},
				title={FedDYN},
				grid=both,
				mark repeat=10,
				mark size=1.4pt,
				thick,
				scale only axis,
				tick label style={font=\small},
				label style={font=\small},
				title style={font=\small}
				]
				\addplot[color=blue, mark=*] table[x index=0, y=FedDYN_base, col sep=comma] {pacsdatac.csv};
				\addplot[color=red, mark=square*] table[x index=0, y=FedDYN_graded, col sep=comma] {pacsdatac.csv};
				\addplot[color=green!60!black, mark=triangle*] table[x index=0, y=FedDYN_random, col sep=comma] {pacsdatac.csv};
			\end{axis}
		\end{tikzpicture}
		&
		\begin{tikzpicture}
			\begin{axis}[
				width=0.43\textwidth,
				height=0.28\textwidth,
				xmin=1, xmax=200,
				ymin=50, ymax=75,
				xlabel=Round,
				ylabel={}, 
				title={FedDC},
				grid=both,
				mark repeat=10,
				mark size=1.4pt,
				thick,
				scale only axis,
				tick label style={font=\small},
				label style={font=\small},
				title style={font=\small}
				]
				\addplot[color=blue, mark=*] table[x index=0, y=FedDC_base, col sep=comma] {pacsdatac.csv};
				\addplot[color=red, mark=square*] table[x index=0, y=FedDC_graded, col sep=comma] {pacsdatac.csv};
				\addplot[color=green!60!black, mark=triangle*] table[x index=0, y=FedDC_random, col sep=comma] {pacsdatac.csv};
			\end{axis}
		\end{tikzpicture}
		\\[6pt]
		
		\multicolumn{2}{c}{
			\begin{tikzpicture}
				\begin{axis}[
					width=0.9\textwidth,     
					height=0.28\textwidth,
					xmin=1, xmax=250,
					ymin=35, ymax=75,
					xlabel=Round,
					ylabel={global Test Accuracy (\%)},
					title={FedProc},
					grid=both,
					mark repeat=10,
					mark size=1.4pt,
					thick,
					scale only axis,
					tick label style={font=\small},
					label style={font=\small},
					title style={font=\small}
					]
					\addplot[color=blue, mark=*] table[x index=0, y=FedProc_base, col sep=comma] {pacsdatac.csv};
					\addplot[color=red, mark=square*] table[x index=0, y=FedProc_graded, col sep=comma] {pacsdatac.csv};
					\addplot[color=green!60!black, mark=triangle*] table[x index=0, y=FedProc_random, col sep=comma] {pacsdatac.csv};
				\end{axis}
			\end{tikzpicture}
		}
	\end{tabular}
	
	\caption{Test accuracy history — Experiment on PACS (C). Each plot shows the base method and two JDFL variants (graded, random).}
	\label{fig:compact-panels_PACSC}
\end{figure*}

\FloatBarrier
\begin{table*}[t]
	\centering
	\tiny
	\caption{Office-Caltech. $E=5$, client\_frac=1, 4 clients. Standalone (standard accuracy) vs.\ Baseline+JDFL (class-only accuracy). Numbers are mean over 3 seeds. Per-domain columns: Amazon / Caltech / DSLR / Webcam.}
	\label{tab:office_e1}
	\resizebox{0.9\textwidth}{!}{%
		\begin{tabular}{lccccc}
			\hline
			Method & Overall & Amazon & Caltech & DSLR & Webcam \\
			\hline
			FedAvg (standalone)            & 73.10 & 84.38 & 60.59 & 87.50 & 70.26 \\
			FedAvg + JDFL (graded)         & 76.51 ({\tiny \textcolor{green}{+3.41}}) & 85.77 & 66.52 & 86.46 & 70.10 \\
			FedAvg + JDFL (random)         & 76.97 ({\tiny \textcolor{green}{+3.87}}) & 86.46 & 65.93 & 87.50 & 80.43 \\
			\hline
			SCAFFOLD (standalone)          & 75.20 & 86.63 & 61.04 & 92.71 & 82.49 \\
			SCAFFOLD + JDFL (graded)       & 77.37 ({\tiny \textcolor{green}{+2.17}}) & 86.81 & 64.59 & 92.71 & 87.01 \\
			SCAFFOLD + JDFL (random)       & 78.61 ({\tiny \textcolor{green}{+3.41}}) & 87.67 & 66.22 & 95.84 & 87.01 \\
			\hline
			FedDyn (standalone)            & 71.72 & 81.77 & 59.41 & 85.42 & 78.53 \\
			FedDyn + JDFL (graded)         & 73.10 ({\tiny \textcolor{green}{+1.38}}) & 83.51 & 60.44 & 86.46 & 80.79 \\
			FedDyn + JDFL (random)         & 73.88 ({\tiny \textcolor{green}{+2.16}}) & 83.51 & 61.78 & 86.46 & 81.92 \\
			\hline
			FedDC (standalone)             & 74.35 & 85.25 & 60.30 & 92.71 & 82.49 \\
			FedDC + JDFL (graded)          & 75.78 ({\tiny \textcolor{green}{+1.43}}) & 84.20 & 63.26 & 90.62 & 88.14 \\
			FedDC + JDFL (random)          & 77.29 ({\tiny \textcolor{green}{+2.94}}) & 85.76 & 64.89 & 93.75 & 88.14 \\
			\hline
			FedProc (standalone)           & 76.82 & 86.11 & 66.37 & 89.58 & 79.66 \\
			FedProc + JDFL (graded)        & 77.95 ({\tiny \textcolor{green}{+1.13}}) & 85.94 & 68.15 & 85.42 & 85.31 \\
			FedProc + JDFL (random)        & 77.82 ({\tiny \textcolor{green}{+1.00}}) & 85.94 & 68.00 & 86.46 & 84.17 \\
			\hline
		\end{tabular}%
	}
\end{table*}

\begin{table*}[t]
	\centering
	\tiny
	\caption{PACS, Setting A. $E=1$, client\_frac=1.0, 10 clients. Standalone vs.\ Baseline+JDFL. Numbers are mean over 3 seeds. Per-domain columns: Art / Cartoon / Photo / Sketch.}
	\label{tab:pacs_e1}
	\resizebox{0.9\textwidth}{!}{%
		\begin{tabular}{lccccc}
			\hline
			Method & Overall & Art & Cartoon & Photo & Sketch \\
			\hline
			FedAvg (standalone)            & 69.40 & 47.64 & 74.33 & 69.66 & 77.69 \\
			FedAvg + JDFL (graded)         & 70.30 ({\tiny \textcolor{green}{+0.90}}) & 49.43 & 75.46 & 72.25 & 77.27 \\
			FedAvg + JDFL (random)         & 71.80 ({\tiny \textcolor{green}{+2.40}}) & 52.52 & 76.31 & 71.86 & 79.13 \\
			\hline
			SCAFFOLD (standalone)          & 70.63 & 47.15 & 76.46 & 70.66 & 79.39 \\
			SCAFFOLD + JDFL (graded)       & 71.97 ({\tiny \textcolor{green}{+1.34}}) & 49.59 & 80.71 & 71.66 & 78.54 \\
			SCAFFOLD + JDFL (random)       & 71.86 ({\tiny \textcolor{green}{+1.23}}) & 47.80 & 78.86 & 74.65 & 79.05 \\
			\hline
			FedDyn (standalone)            & 69.00 & 41.79 & 73.62 & 71.46 & 79.39 \\
			FedDyn + JDFL (graded)         & 70.63 ({\tiny \textcolor{green}{+1.63}}) & 45.85 & 76.59 & 73.05 & 78.88 \\
			FedDyn + JDFL (random)         & 69.57 ({\tiny \textcolor{green}{+0.57}}) & 45.20 & 73.76 & 67.87 & 80.49 \\
			\hline
			FedDC (standalone)             & 68.27 & 42.77 & 75.18 & 67.07 & 77.95 \\
			FedDC + JDFL (graded)          & 70.50 ({\tiny \textcolor{green}{+2.23}}) & 45.37 & 78.44 & 68.86 & 79.56 \\
			FedDC + JDFL (random)          & 69.27 ({\tiny \textcolor{green}{+1.00}}) & 44.55 & 75.18 & 68.66 & 78.88 \\
			\hline
			FedProc (standalone)           & 71.03 & 49.27 & 75.04 & 75.85 & 77.95 \\
			FedProc + JDFL (graded)        & 72.47 ({\tiny \textcolor{green}{+1.44}}) & 51.22 & 77.73 & 75.45 & 79.13 \\
			FedProc + JDFL (random)        & 72.57 ({\tiny \textcolor{green}{+1.54}}) & 52.68 & 77.02 & 73.45 & 79.89 \\
			\hline
		\end{tabular}%
	}
\end{table*}

\begin{table*}[t]
	\centering
	\tiny
	\caption{PACS, Setting B. $E=5$, client\_frac=0.5, 10 clients. Standalone vs.\ Baseline+JDFL. Numbers are mean over 3 seeds. Per-domain columns: Photo / Art / Cartoon / Sketch.}
	\label{tab:pacs_e2}
	\resizebox{0.9\textwidth}{!}{%
		\begin{tabular}{lccccc}
			\hline
			Method & Overall & Art & Cartoon & Photo & Sketch \\
			\hline
			FedAvg (standalone)            & 67.90 & 45.04 & 72.20 & 69.26 & 76.68 \\
			FedAvg + JDFL (graded)         & 69.87 ({\tiny \textcolor{green}{+1.97}}) & 47.32 & 75.32 & 71.06 & 77.86 \\
			FedAvg + JDFL (random)         & 70.53 ({\tiny \textcolor{green}{+2.63}}) & 48.29 & 74.75 & 73.06 & 78.54 \\
			\hline
			SCAFFOLD (standalone)          & 69.83 & 46.99 & 75.17 & 70.46 & 78.29 \\
			SCAFFOLD + JDFL (graded)       & 71.83 ({\tiny \textcolor{green}{+2.00}}) & 49.10 & 78.16 & 72.66 & 79.56 \\
			SCAFFOLD + JDFL (random)       & 72.80 ({\tiny \textcolor{green}{+2.97}}) & 49.11 & 79.15 & 74.45 & 80.66 \\
			\hline
			FedDyn (standalone)            & 67.47 & 43.74 & 73.33 & 66.87 & 76.59 \\
			FedDyn + JDFL (graded)         & 68.40 ({\tiny \textcolor{green}{+0.93}}) & 45.86 & 74.18 & 67.86 & 76.93 \\
			FedDyn + JDFL (random)         & 67.60 ({\tiny \textcolor{green}{+0.13}}) & 43.41 & 71.63 & 69.06 & 77.27 \\
			\hline
			FedDC (standalone)             & 69.80 & 46.34 & 75.74 & 70.46 & 78.20 \\
			FedDC + JDFL (graded)          & 71.43 ({\tiny \textcolor{green}{+1.63}}) & 48.62 & 78.72 & 70.86 & 79.22 \\
			FedDC + JDFL (random)          & 70.03 ({\tiny \textcolor{green}{+0.23}}) & 42.93 & 76.46 & 69.86 & 80.41 \\
			\hline
			FedProc (standalone)           & 69.03 & 46.99 & 74.61 & 71.46 & 76.17 \\
			FedProc + JDFL (graded)        & 70.57 ({\tiny \textcolor{green}{+1.54}}) & 49.27 & 76.31 & 68.66 & 79.05 \\
			FedProc + JDFL (random)        & 70.17 ({\tiny \textcolor{green}{+1.14}}) & 50.57 & 73.34 & 70.46 & 78.37 \\
			\hline
		\end{tabular}%
	}
\end{table*}

\begin{table*}[t]
	\centering
	\tiny
	\caption{PACS, Setting C. $E=3$, client\_frac=0.5, 40 clients. Standalone vs.\ Baseline+JDFL. Numbers are mean over 3 seeds. Per-domain columns: Photo / Art / Cartoon / Sketch.}
	\label{tab:pacs_e3}
	\resizebox{0.9\textwidth}{!}{%
		\begin{tabular}{lccccc}
			\hline
			Method & Overall & Art & Cartoon & Photo & Sketch \\
			\hline
			FedAvg (standalone)            & 68.53 & 46.13 & 72.90 & 69.87 & 76.93 \\
			FedAvg + JDFL (graded)         & 69.23 ({\tiny \textcolor{green}{+0.70}}) & 50.08 & 72.20 & 69.66 & 77.27 \\
			FedAvg + JDFL (random)         & 70.07 ({\tiny \textcolor{green}{+1.54}}) & 49.76 & 74.18 & 69.26 & 78.54 \\
			\hline
			SCAFFOLD (standalone)          & 69.27 & 45.37 & 76.60 & 70.46 & 76.84 \\
			SCAFFOLD + JDFL (graded)       & 70.47 ({\tiny \textcolor{green}{+1.20}}) & 48.46 & 77.59 & 71.85 & 77.10 \\
			SCAFFOLD + JDFL (random)       & 70.13 ({\tiny \textcolor{green}{+0.86}}) & 46.51 & 77.30 & 71.86 & 77.44 \\
			\hline
			FedDyn (standalone)            & 67.43 & 40.00 & 71.91 & 68.46 & 78.63 \\
			FedDyn + JDFL (graded)         & 69.50 ({\tiny \textcolor{green}{+2.07}}) & 46.83 & 75.74 & 69.66 & 77.52 \\
			FedDyn + JDFL (random)         & 68.67 ({\tiny \textcolor{green}{+1.24}}) & 43.57 & 74.89 & 67.65 & 78.45 \\
			\hline
			FedDC (standalone)             & 69.30 & 43.47 & 76.31 & 70.26 & 77.53 \\
			FedDC + JDFL (graded)          & 69.77 ({\tiny \textcolor{green}{+0.47}}) & 44.39 & 76.74 & 70.06 & 78.71 \\
			FedDC + JDFL (random)          & 69.37 ({\tiny \textcolor{green}{+0.07}}) & 43.09 & 76.74 & 69.66 & 78.54 \\
			\hline
			FedProc (standalone)           & 69.07 & 46.82 & 71.77 & 70.66 & 78.20 \\
			FedProc + JDFL (graded)        & 69.73 ({\tiny \textcolor{green}{+0.66}}) & 49.43 & 72.34 & 71.26 & 78.12 \\
			FedProc + JDFL (random)        & 69.77 ({\tiny \textcolor{green}{+0.70}}) & 51.71 & 71.63 & 69.66 & 78.12 \\
			\hline
		\end{tabular}%
	}
\end{table*}